\documentclass[letterpaper, 10 pt, conference]{ieeeconf}
\IEEEoverridecommandlockouts    
\usepackage{amssymb}
\usepackage{amsmath}
\usepackage{xcolor}

\usepackage{flushend}
\usepackage{multirow}

\usepackage{makecell}
\usepackage{adjustbox}
\usepackage{lscape} 

\usepackage{lineno}

\usepackage{graphics}           
\usepackage{times}              
\usepackage{amsmath}            
\usepackage{amssymb}            
\usepackage{graphicx}
\usepackage{algorithm}
\usepackage[noend]{algpseudocode}
\usepackage{booktabs}
\usepackage{color}
\usepackage{listings}
\usepackage{subfiles}
\usepackage{hyperref}
\usepackage[nocompress]{cite} 
\definecolor{instructioncolor}{rgb}{.5,.5,.5}

\usepackage[font=small]{caption}

\def\secref#1{Sec.~\ref{#1}}
\def\figref#1{Fig.~\ref{#1}}
\def\tabref#1{Tab.~\ref{#1}}
\def\eqref#1{Eq.~(\ref{#1})}

\makeatletter
\usepackage{xspace}
\DeclareRobustCommand\onedot{\futurelet\@let@token\@onedot}
\def\@onedot{\ifx\@let@token.\else.\null\fi\xspace}
\def\eg{e.g\onedot} 
\def\ie{i.e\onedot}

\def\etal{{et al}\onedot}
\def\ps@pprintTitle{%
  \let\@oddhead\@empty
  \let\@evenhead\@empty
  \let\@oddfoot\@empty
  \let\@evenfoot\@oddfoot
}
\makeatother

\usepackage{array}
\newcolumntype{L}[1]{>{\raggedright\let\newline\\\arraybackslash\hspace{0pt}}m{#1}}
\newcolumntype{C}[1]{>{\centering\let\newline\\\arraybackslash\hspace{0pt}}m{#1}}
\newcolumntype{R}[1]{>{\raggedleft\let\newline\\\arraybackslash\hspace{0pt}}m{#1}}

\def\argmax{\mathop{\rm argmax}}

\def\argmin{\mathop{\rm argmin}}

\newcommand{\new}[1]{\textcolor{black}{#1}}
\newcommand{\nnew}[1]{\textcolor{black}{#1}}
\newcommand{\minitab}[2][l]{\begin{tabular}{#1}#2\end{tabular}}

\definecolor{graph_yellow}{rgb}{0.90588,0.8313725,0.6078431}
\definecolor{graph_red}{rgb}{0.811764706,0.560784314,0.525490196}
\definecolor{graph_green}{rgb}{0.647058824, 0.803921569, 0.776470588}

\definecolor{circle_fuchsia}{rgb}{1.0, 0.27843, 0.98039}
\definecolor{circle_orange}{rgb}{1.0, 0.77647, 0.22745}

\def\etalcite#1{\etal~\cite{#1}}

\usepackage{fancyhdr}
\fancypagestyle{arxivhdr}
{
   \fancyhf{}
   \setlength{\headheight}{15pt} 
\fancyfoot[C]{This paper has been accepted for publication in Computers and Electronics in Agriculture\\ \url{https://doi.org/10.1016/j.compag.2026.111723}}
\fancyhead[C]{\footnotesize Please cite this paper as:\\
D. Fusaro, F. Magistri, J. Behley, A. Pretto, C. Stachniss, "Horticultural temporal fruit monitoring via 3D instance segmentation and re-identification using colored point clouds," in Computers and Electronics in Agriculture, Volume 247,  2026, \url{https://doi.org/10.1016/j.compag.2026.111723}}
}

\title{Horticultural Temporal Fruit Monitoring via 3D Instance Segmentation and Re-Identification using Colored Point Clouds}

\author{Daniel Fusaro \and Federico Magistri \and Jens Behley \and Alberto Pretto \and Cyrill Stachniss
 \thanks{D. Fusaro and A. Pretto are with the University of Padua, Italy. \mbox{F. Magistri}, J. Behley, and C. Stachniss are with the Center for Robotics, University of Bonn, Germany. () C. Stachniss is additionally with the Department of Engineering Science at the University of Oxford, UK, and with the Lamarr Institute for Machine Learning and Artificial Intelligence, Germany.}%
}

\begin{document}
\maketitle

\thispagestyle{arxivhdr}
\pagestyle{empty}

\begin{abstract}
Accurate and consistent fruit monitoring over time is a key step toward automated agricultural production systems.
However, this task is inherently difficult due to variations in fruit size, shape, occlusion, orientation, and the dynamic nature of orchards where fruits may appear or disappear between observations.
In this article, we propose a novel method for fruit instance segmentation and re-identification on 3D terrestrial point clouds collected over time.
Our approach directly operates on dense colored point clouds, capturing fine-grained 3D spatial detail.
We segment individual fruits using a learning-based instance segmentation method applied directly to the point cloud.
For each segmented fruit, we extract a compact and discriminative descriptor using a 3D sparse convolutional neural network.
To track fruits across different times, we introduce an attention-based matching network that associates fruits with their counterparts from previous sessions.
Matching is performed using a probabilistic assignment scheme, selecting the most likely associations across time.
We evaluate our approach on real-world datasets of strawberries and apples, demonstrating that it outperforms existing methods in both instance segmentation and temporal re-identification, enabling robust and precise fruit monitoring across complex and dynamic orchard environments.

\end{abstract}









\section{Introduction}
\label{sec:intro}


The challenge of meeting the growing demand for food requires advances in agricultural practices with a focus on efficiency and sustainability.
Autonomous robots offer new possibilities to automate labor-intensive tasks such as crop monitoring and management.
Such systems have the potential to \nnew{improve} agricultural production systems and can enable continuous and large-scale monitoring~\cite{duckett2018arxiv, walter2018nas}.
In particular, these technologies support phenotyping, the process of evaluating plant characteristics by providing precise, high-throughput assessments and surpassing the limitations of traditional manual methods~\cite{fiorani2013arpb, watt2020arpb}.
This shift towards automated phenotyping represents a critical step forward in optimizing crop selection and improving crop yield.
Temporal matching, or fruit re-identification, combined with accurate instance segmentation, enables tracking the development of single fruits over time, supporting the analysis of growth patterns and estimation of maturation rates.
To enable such monitoring, the perception system is essential \nnew{to ensure a} reliable visual association over time.

\begin{figure}[ht]
  \centering
  \includegraphics[width=1.0\linewidth]{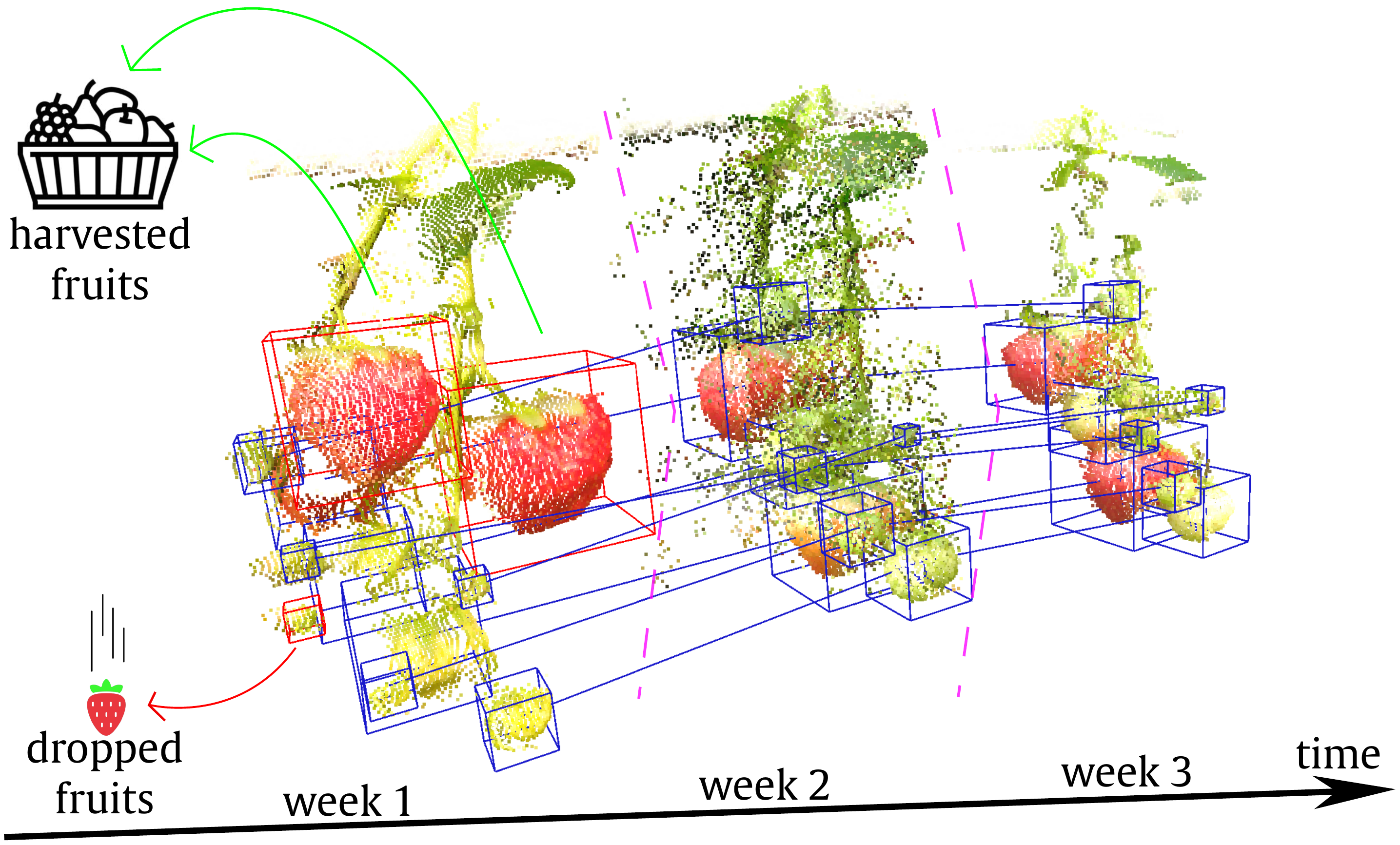}
  \caption{\new{Fruit re-identification on three point clouds acquired at three different points in time.
  Fruit instances are first segmented using an instance segmentation method, then they are temporally matched with fruit instances of a previous data collection (\eg, matching fruits recorded in week~$t$ with fruits from week~$t+1$).
  Blue bounding boxes (\textcolor{blue}{\rule[-0.07ex]{0.2cm}{1.6ex}}) indicate matched fruit instances, while red bounding boxes (\textcolor{red}{\rule[-0.07ex]{0.2cm}{1.6ex}}) indicate unmatched fruits (\eg,~harvested, dropped or newly appeared).}}
  \label{fig:motivation}
\end{figure} 

In this article, we aim to recognize and track object instances, specifically fruits, using real, colored point clouds acquired by a high-resolution LiDAR scanner.
The goal is to identify, segment, and temporally match individual fruits at different points in time within the 3D space.
2D image-based approaches~\cite{perez-borrero2020cea, ganesh2019ifac, gonzalez2019ieeeaccess, genemola2020cea, weikuan2021cea} lack the 3D structural, depth, and spatial information offered by the point clouds.
\nnew{In contrast}, point clouds lack a regular grid structure, making it difficult to apply traditional image processing techniques.
The challenge is to process these sparse and irregular data to detect and associate individual fruits, which may vary in size, shape, orientation, and occlusion levels and may be harvested or newly grown.

Once fruit instances have been segmented, the temporal re-identification task involves recognizing and matching the same fruit instances across different point clouds captured at different points in time or from different viewpoints.
This task shares challenges with object re-identification and temporal association in dynamic 3D scenes, and is conceptually related to loop closures in SLAM~\cite{stachniss2016handbook-slamchapter} and visual place recognition~\cite{vysotska2019ral} systems, which typically rely on uniquely identifiable landmarks.
In the context of fruit re-identification, there are no unique traits that make fruits easily distinguishable.
As depicted in~\figref{fig:motivation}, fruits can be very similar, tightly packed, and their pose can change significantly over time, causing trivial solutions based on relative position to fail.


The main contribution of this article is a novel method for accurately performing fruit instance segmentation and re-identification on terrestrial point clouds captured at different points in time, based on a learned descriptor encoder and an attentive matcher.
We exploit dense high-precision point clouds recorded with a high-precision Faro laser scanner, which enables fine-grained spatial detail in 3D object representation.
\nnew{Although} these sensors are not commonly employed in robotic applications, their capability to scan detailed environments has recently garnered attention, leading to their integration into robotic systems~\cite{rodriguez2024fps}.
We segment fruits using a learning-based instance segmentation, which infers fruit instances directly from the point cloud.
Each fruit is then processed by a 3D sparse convolutional neural network to extract a compact, discriminative descriptor.
Then, we match each fruit with its corresponding instance from a previous data collection by using their descriptors and an attention-based matching module.
To handle the possibility of a no-match scenario, where a query fruit is identified as a new instance, we represent it using a specific descriptor.
We then predict a probability distribution over the candidate fruits from a previous data collection, explicitly including a no-match class.
Each query fruit is subsequently matched with the previous instance that has the highest predicted probability, using a greedy assignment to determine associations.

In sum, we make two key claims:
(i) our approach is able to identify fruits in point clouds using an instance segmentation method;
(ii) it outperforms baseline approaches on the instance segmentation and re-identification tasks using real-world 3D data. 
These claims \nnew{are supported} by the paper and our experimental evaluation.
Using 3D data and sparse convolution neural networks, our method significantly enhances the effectiveness and scalability of object-level segmentation and temporal association in sparse point clouds.
It outperforms baseline approaches and offers new capabilities for automated monitoring in dynamic 3D environments, contributing to the broader field of 3D pattern recognition.

The implementation of our fruit matching method is publicly available at \href{https://github.com/PRBonn/IRIS3D}{https://github.com/PRBonn/IRIS3D}.

\section{Related Works} \label{sec:related}  
Our work intersects several research areas: instance segmentation, plant phenotyping, and temporal object matching. We begin by reviewing 2D and 3D instance segmentation methods, including both general-purpose approaches and those specifically designed for agricultural applications. We then discuss image-based and 3D plant phenotyping techniques focused on extracting structural traits from crops. Finally, we review temporal matching techniques for tracking objects over time, a critical task in agricultural 3D data analysis.

\textit{Instance segmentation on 2D images}:
General instance segmentation, the task of segmenting individual objects within a scene, has been extensively studied in the context of 2D images. In this domain, the \mbox{Mask R-CNN}~\cite{he2017iccv-mr} architecture has emerged as one of the most popular methods. It is a deep learning architecture that extends the object detection process by adding a parallel branch that predicts segmentation masks, allowing precise localization of object boundaries. 
It has been applied to several fruit instance segmentation tasks on images~\cite{perez-borrero2020cea, ganesh2019ifac, gonzalez2019ieeeaccess, genemola2020cea}.
Unlike the standard Mask R-CNN, Weikuan~\etalcite{weikuan2021cea} implement an anchor-free inference pipeline intended to make the model more stable and easily applicable to other green fruits without hyper-parameter tuning.

\textit{\new{Instance segmentation on RGB-D images: }}
\new{While 2D image-based instance segmentation methods have shown promising results, they often struggle with occlusions and overlapping objects, which are common in agricultural environments.
To address these challenges, some approaches have incorporated depth information from RGB-D images.
Ge~\etalcite{ge2019ifac} adopts Mask R-CNN~\cite{he2017iccv-mr} to segment and localize fruits in RGB-D images.
Kang~\etalcite{kang2020cea} propose DaSNet-v2, a multi-task network that jointly performs detection and instance segmentation on fruits, and semantic segmentation on branches, in RGB-D images collected in apple orchards.
It applies feature pyramid networks and atrous spatial pyramid pooling to effectively capture multi-scale features and context information.
Tang~\etalcite{tang2024fsfs} propose a high-precision apple instance segmentation method based on an improved SOLOv2~\cite{wang2020nipsjournel} and EfficientNet~\cite{tan2019icml} backbone using RGB-D images.
In scenarios involving overlapping or occluded apples, authors apply a lightweight spatial attention module to improve segmentation accuracy.
Magistri~\etalcite{magistri2024ral} exploit shape completion and differentiable rendering techniques to estimate the 3D shape of a target fruit together with its pose even under strong occlusions from a single \mbox{RGB-D} image.
RGB-D images provide additional spatial information that can help disambiguate overlapping objects and improve segmentation accuracy in complex scenes, but they still lack the full 3D structural detail that point clouds can offer.
The point clouds generated from~RGB-D images are typically sparse and noisy, which can limit the effectiveness of~3D instance segmentation methods.
}

\textit{Instance segmentation on 3D point clouds}:
The task of instance segmentation becomes significantly more challenging when applied to 3D data such as point clouds obtained using LiDAR sensors. 
Most methods are based on deep learning and rely on the use of 3D convolutional neural networks (CNNs) to learn features from the point cloud data.
Although they can be trained end-to-end, they often require a large amount of labeled data to achieve good performance. Data-augmentation techniques~\cite{zhu2024pr} can be used to artificially increase the size of the training dataset, but the lack of large-scale datasets for 3D instance segmentation, especially for agricultural robotics, remains a challenge.
Most neural network-based approaches for instance segmentation on point clouds~\cite{choy2019cvpr, zhu2022tpami, schult2023icra, marcuzzi2023ral, robert20243dv, shin2024cvpr, xiao2025ijcv, kolodiazhnyi2024cvpr, hong2021cvpr, xiang2023isprs} voxelize the 3D point cloud to preserve topological relations and use sparse convolutions~\cite{graham2018cvpr} to reduce the memory consumption.
Schult~\etalcite{schult2023icra} proposed Mask3D. Based on transformer decoders~\cite{vaswani2017neurips}, it leverages learned instance queries together with point-wise features to directly predict semantic instance masks, eliminating the need for handcrafted voting schemes or grouping heuristics. A multi-scale, sparse convolution-based backbone, along with a query refinement mechanism, contributed to its state-of-the-art performance on several benchmarks at the time of its introduction.
Marcuzzi~\etalcite{marcuzzi2023ral} proposed MaskPLS, tailored for autonomous driving scenarios. Similarly to Mask3D, their approach incorporates a multi-scale sparse convolutional backbone, a query refinement strategy, and transformer decoders. The inclusion of intermediate losses and other architectural choices further enhanced \nnew{the effectiveness of the method}.
Superpoint transformer, introduced by Robert~\etalcite{robert20243dv}, is a fast and light-weight state-of-the-art approach for indoor and outdoor point clouds. It is based on a pre-processing step using handcrafted features, multi-scale superpoints generation, and graph-attention convolutional networks.
Spherical Mask, suggested by Shin~\etalcite{shin2024cvpr}, is also built on a 3D backbone that utilizes sparse convolutions. A voting module is used to generate instance queries, after which a 3D polygon, represented by points and rays, is estimated for each query. To enable fine-grained clustering, the method predicts point-wise offsets that adjust the spatial distribution of points around each proposal.
Despite these advances, to the best of our knowledge, there is no work that has specifically addressed 3D fruit instance segmentation directly on 3D point clouds.
Closely related, Kang~\etalcite{kang2022cea} fuse point clouds and images to perform fruit localization using a single-stage instance segmentation network.

\textit{Image and 3D plant phenotyping}:
Image-based phenotyping \nnew{for automated plant monitoring} has become increasingly important in numerous agricultural settings. 
Computer vision, aided by deep learning techniques, has been applied to different phenotyping-oriented agricultural contexts~\cite{rodriguez2020prl, liu2024pr, chu2021prl, cardellicchio2024prl, kierdorf2021fai, nuske2011iros, halstead2018ral, smitt2021icra, blok2021biosyseng, gomez2021ral}.
Although image-based phenotyping has garnered considerable attention, there are relatively few studies that proposed methods for plant phenotyping using 3D data.
Hao~\etalcite{hao2024be} analyzes and evaluates the degree of wilting of cotton varieties in point clouds. 
Boogaard~\etalcite{boogaard2023be} investigates \nnew{the problem of measurement of internode length} in cucumbers by comparing estimates from 3D point clouds with estimates from images.
A key factor contributing to the gap between 2D and 3D analysis lies \nnew{in the limitations of the sensor}. For example, conventional sensors used on robots, such as 3D LiDARs and \mbox{RGB-D} cameras, typically offer a 3D spatial resolution that is insufficiently detailed for agricultural environments.
One potential solution is to utilize high-precision laser scanners to generate dense, colored, and highly accurate point clouds.
Such data have also been used in the past. For example, Rodriguez-Sanchez~\etalcite{rodriguez2024ground} demonstrate the use of a ground robot to automate \nnew{the acquisition of data} from terrestrial laser scanners in a breeding field.

\textit{Temporal matching}:
There \nnew{are} also a limited number of studies that focus on temporal fruit matching, \ie, the problem of finding fruit correspondences over time.
Chebrolu~\etalcite{chebrolu2021plosone} exploit a skeletal structure of the complete plant to compute correspondences between the same plant over two weeks to estimate leaf growth parameters.
Riccardi~\etalcite{riccardi2023icra} propose a histogram descriptor that uses Euclidean distance measurements between neighboring fruits. For a given target fruit, their method divides the surrounding 3D space into angular sectors and counts the number of fruits within each sector to build a descriptor that can be used for temporal matching.
Lobefaro~\etalcite{lobefaro2023iros} investigate the problem of 4D data association of growing pepper plants in a greenhouse by combining 3D \mbox{RGB-D} SLAM to build local plant models and visual place recognition to create correspondences over time. 
The same authors, in a follow-up work~\cite{lobefaro2024iros}, employ deep-learning-based feature descriptors and geometric information to obtain matches between 3D points and track the evolution of plant traits’ over time.\\

In contrast to \nnew{previous} work, we rely on MinkPanoptic (\new{a core component module} of~MaskPLS~\cite{marcuzzi2023ral}) for our instance segmentation method. Based on such a segmentation, we can address different downstream tasks, such as plant and fruit phenotyping.
We use a learned descriptor, based on 3D sparse convolution, to automatically learn and extract relevant features from raw data through training, allowing it to adapt to complex patterns and variations in the data. Also, our learned descriptor can adapt to new and unseen data without requiring explicit re-engineering of feature extraction methods. We perform descriptor matching leveraging an attention-based network that predicts a probability distribution over the pool of candidate matchings, also considering the no-match case.

\section{Our Approach} \label{sec:main}

\begin{figure*}[t]
  \centering
  \includegraphics[width=\linewidth]{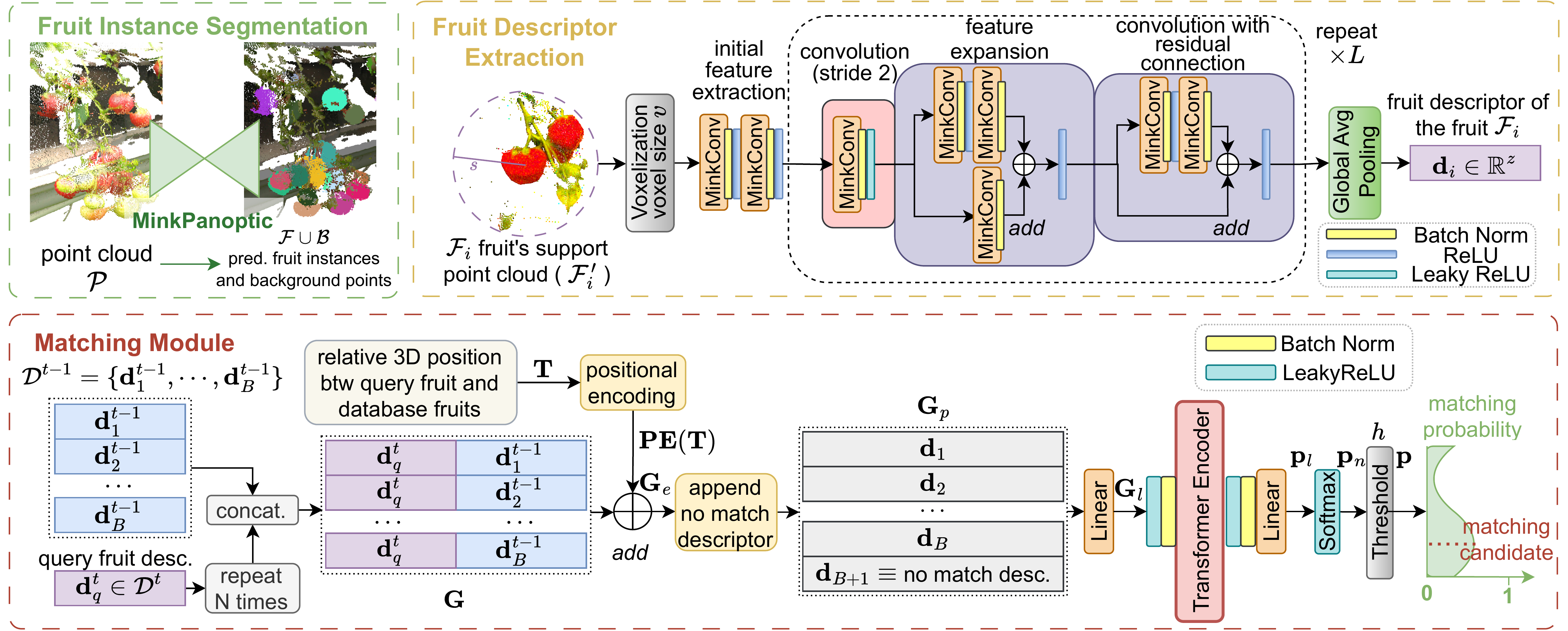}
  \caption{Pipeline of our approach. Fruit instance segmentation provides $~K$ fruit instance masks $\mathcal{F}=~\{{\mathcal{F}_i}\}^{K}_{i=1}$ using a colored point cloud $\mathcal{P}$.
                The fruit descriptor extraction module processes a fruit's support point cloud, $\mathcal{F}'_i$, and computes the fruit descriptor, $\mathbf{d}_i$.
                It initially voxelizes the input point cloud, then a MinkowskiNet~\cite{choy2019cvpr} encoder processes the voxelized point cloud by leveraging sparse 3D convolutions.
                A final global average pooling aggregates the features of all voxels to compute the descriptor.
                The matching module matches a query descriptor $\mathbf{d}_i$ with a set of descriptors $\mathcal{D}^{t-1}$ of fruits that belong to a different point in time.}
  \label{fig:FDM-arch}
\end{figure*}


We aim to track fruits using real 3D data acquired by a high-resolution LiDAR scanner. 
We present a novel method for accurately performing fruit instance segmentation and re-identification on point clouds captured at different points in time.

\subsection{Fruit Instance Segmentation Module} \label{sec:instsegm}
Consider a colored point cloud $\mathcal{P}$ in which each point is given by its 3D position $\mathbf{p}_i \in \mathbb{R}^3$ and its RGB color $\mathbf{k}_i$.
We want to segment $\mathcal{P}$ into a set $\mathcal{F}=~\{{\mathcal{F}_i}\}^{K}_{i=1}$ of $~K$ fruit instances and a set $\mathcal{B}$ of background points.
Each $\mathcal{F}_i \subseteq \mathcal{P}$ can be described by the pair $(\mathbf{c}_i, r_i)$, where $\mathbf{c}_i \in \mathbb{R}^3$ is the center of the fruit, given by:
\begin{equation}
\mathbf{c}_i = \frac{1}{|\mathcal{F}_i|} \sum_{\mathbf{p}_j \in \mathcal{F}_i}{\mathbf{p}_j},
\end{equation}
and $r_i$ is the radius of the smallest sphere, centered at $\mathbf{c}_i$, containing all the points of the fruit.
Note that $\mathbf{c}_i$ does not necessarily correspond to a point in the point cloud.

We use MinkPanoptic~\cite{marcuzzi2023ral}, a learning-based instance segmentation method, to obtain fruit instances in a point cloud. We selected this method empirically, as it consistently outperformed others in our experiments, particularly in the context of the task presented in this article, where training data are limited (see~\secref{sec:exp}).

Let $\mathcal{P}_f \subseteq \mathcal{P}$ be the subset of $\mathcal{P}$ that was labeled as fruit.
Then, for each point in $\mathcal{P}_f$, the method predicts a 3D vector representing the offset of the point from the center of the fruit instance to which it belongs.
We add the predicted offsets to $\mathcal{P}_f$ to obtain the point cloud $~\mathcal{P}_o$.
We determine fruit instances $\mathcal{F}$ by clustering $~\mathcal{P}_o$ using the mean shift~\cite{comaniciu2002pami} algorithm.
Clustering-based methods generally achieve better performance than end-to-end instance segmentation approaches \cite{xiang2023isprs}.

\subsection{Fruit Descriptor Extraction Module} \label{sec:method-fdem}
The fruit descriptor extraction module processes fruit instances obtained with the instance segmentation method described in \secref{sec:instsegm} and computes their descriptors.
Given a fruit $\mathcal{F}_i$ and its center $\mathbf{c}_i$, we define with $\mathcal{S}_i \subseteq \mathcal{P}$ the support of $\mathcal{F}_i$ with fixed radius $s$:
\begin{equation}
\mathcal{S}_i = \{ (\mathbf{p}, \mathbf{k}) \in \mathcal{P} \mid \|\mathbf{p} - \mathbf{c}_i\|_2 \leq s \}.
\end{equation}

The usage of $\mathcal{S}_i$ instead of $\mathcal{F}_i$ enables the \nnew{extraction of the descriptor to also account for} the surroundings of the fruit~$\mathcal{F}_i$.
We want the descriptor to be as discriminative as possible between different fruits, but as similar as possible for the point clouds of the same fruit taken at different points in time.
In addition, the descriptor should be robust to rotation, shape, and color variation.
Fruits often change their orientation during growth, but their shape also changes substantially. 
The color changes both due to the growth of the fruit and due to the greenhouse's internal light change (position of the sensor, weather conditions, humidity, artificial light, etc.).


On top of~\figref{fig:FDM-arch}, the architecture of the fruit descriptor extraction module is depicted.
First, the fruit support point cloud~$\mathcal{S}_i$~is voxelized using a fixed voxel size~$v$.
Then, a MinkowskiNet~\cite{choy2019cvpr} encoder processes the voxelized point cloud by leveraging sparse 3D convolutions, capturing multi-scale features through a deep hierarchical structure.

The encoder is composed of multiple stages that progressively downsample the input spatial dimensions while increasing the number of feature channels.
It begins with two sparse 3D convolutional layers, each followed by batch normalization (BN)~\cite{ioffe2015icml} and a ReLU, responsible for the initial feature extraction from the input sparse tensor.
Then, $L$~identical blocks sequentially process the sparse tensor.

Each block consists of three parts.
The first part performs downsampling (\nnew{through} a convolution with a stride of~2), reducing spatial resolution while expanding the receptive field.
The second part comprises a main convolutional path that applies two sequential 3D sparse convolutions with a~$3\times 3$~kernel, BN, and ReLU activation, expanding the feature channels, and a secondary path consisting of a single convolution with~$1\times 1$ kernel that matches the dimensions of the input to the output, followed by BN.
The third part is very similar to the second one, except that the secondary path consists of a simple residual connection.

After the $L$ identical blocks, we apply a global average pooling, which aggregates the features across all the voxels, obtaining the fruit's descriptor~$\mathbf{d}_i \in \mathbb{R}^z$.

By computing the descriptors of all fruits~$\mathcal{F}$, we build a set of descriptors~$\mathcal{D} = \{\mathbf{d}_1,\dots,\mathbf{d}_{|\mathcal{F}|}\}$ where~$\mathbf{d}_i$ represents the descriptor of fruit~$\mathcal{F}_i$.

\subsection{Fruit Descriptors Matching Module}
The fruit descriptors matching module operates on sets of descriptors of fruits that belong to different time steps.
Let~$\mathcal{D}^t$ be the set of descriptors of all fruits segmented at time~$t$, \ie, $\mathcal{F}^t = \{\mathcal{F}_1^t, \dots, \mathcal{F}_A^{t}\}$, and $\mathcal{D}^{t-1}$ the set of descriptors of all fruits segmented at time~${t-1}$, \ie, $\mathcal{F}^{t-1} = \{\mathcal{F}_1^{t-1}, \dots, \mathcal{F}_B^{t-1}\}$.

We want to match the descriptors in~$\mathcal{D}^{t}$ with the corresponding descriptors in~$\mathcal{D}^{t-1}$. Moreover, we also want to label new or simply not visible fruits descriptors with a no-match label,~$\varnothing$.
Formally, we want to compute the association vector~$\mathbf{y} = [y_1, \dots, y_A]^\top$ with~$y_i \in \{ \varnothing, 1, \dots, B \}$.

Let~$\mathcal{F}_q^t \in \mathcal{F}^{t}$ be a query fruit point cloud at time~$t$ and~$\mathbf{d}_q^t \in \mathcal{D}^{t}$ be its descriptor.
To infer~$y_q$, the matching module computes the probability distribution $p(\mathbf{y} \mid \mathcal{D}^t, \mathcal{D}^{t-1})$. 
The matching module processes the descriptor $\mathbf{d}_q^t$ and the sets of descriptors~$\mathcal{D}^{t}$ and~$\mathcal{D}^{t-1}$.
It also exploits the relative position $\mathbf{T} \in \mathbb{R}^{B \times 3}$ between the query fruit and the fruits in~$\mathcal{F}^{t-1}$.

\figref{fig:FDM-arch} below depicts the architecture of the matching module.
First,~$\mathbf{d}_q^t$ is concatenated to each descriptor in~$\mathcal{D}^{t-1}$, obtaining the matrix~$\mathbf{G} \in \mathbb{R}^{B \times 2z}$
To focus the module on the neighbors of the query fruit, we add a fixed positional encoding~$\mathbf{PE}$ to~$\mathbf{G}$ obtaining~$\mathbf{G}_e$:
\begin{equation}
\mathbf{G}_e = \mathbf{G} + \mathbf{PE}(\mathbf{T}).
\end{equation}

We compute~$\mathbf{PE}:\mathbb{R}^3 \mapsto \mathbb{R}^z$ as a function of the relative position matrix~$\mathbf{T}$ applied to each row as in Mildenhall~\etalcite{mildenhall2020eccv}, \ie, we apply it to each individual coordinate.

To let the matching module account for the no-match case, we artificially create a new row in~$\mathbf{G}_e$ representing the no-match case. The new element is a descriptor with all zero elements and is appended to~$\mathbf{G}_e$ as the last row, obtaining the matrix~$\mathbf{G}_p \in \mathbb{R}^{(B+1) \times 2z}$.

A linear layer, followed by BN and leaky ReLU, maps~$\mathbf{G}_p$ to~$\mathbf{G}_l \in \mathbb{R}^{(B+1) \times l}$, and a transformer encoder layer~\cite{vaswani2017neurips}, followed by BN and leaky ReLU, processes~$\mathbf{G}_l$ to obtain~$\mathbf{G'}_l \in \mathbb{R}^{(B+1) \times l}$.
A final linear layer maps~$\mathbf{G'}_l$ to the prediction logits vector~$\mathbf{p}_l \in \mathbb{R}^{B+1}$, which are then normalized to a probability distribution~$\mathbf{p}_n$ by applying a softmax.
To impose the assumption that fruits do not move, between scans, more than~$h \in \mathbb{R}$ meters, we mask~$\mathbf{p}_n$ by setting the probability of fruits whose center is at a distance greater than~$h$ to~$0$, obtaining~$\mathbf{p}$.

The predicted matching fruit's index~$y_q$ is given by:
\begin{equation}
y_q = \begin{cases} i, & \text{if } i<B+1  \\ \varnothing  & \text{if } i=B+1 \end{cases},
~\text{with}~~i=\argmax_{i}{(\mathbf{p})}.
\end{equation}

Using the prediction~$y_i$ for each fruit in~$\mathcal{F}_i^{t}$, we can associate each fruit in~$\mathcal{F}_i^{t}$ with fruits in~$\mathcal{F}_i^{t-1}$ with the vector~$\mathbf{y}$.

\subsection{Batch Matching}
When matching more than one fruit, we post-process~$\mathbf{y}$ using a greedy matching algorithm.
Specifically, let~$\mathbf{H^*}\in~\mathbb{R}^{A \times (B+1)}$ be the matrix representing all the probability distributions predicted with the matching module, associating all the fruits in $\mathcal{F}^t$ with those in $\mathcal{F}^{t-1}$.
The element~$\mathbf{H^*}_{ij}$ in the row $i$ and column $j \leq B$ represents the probability that the fruit~$\mathcal{F}_{i}^t$ matches with~$\mathcal{F}_j^{t-1}$. The last column of~$\mathbf{H^*}$ encodes the no-match probability of each fruit in $\mathcal{F}^t$.
In addition, each row~$i$ of~$\mathbf{H^*}$ corresponds to the vector~$\mathbf{p}$ computed for~$\mathcal{F}_i^t$.

We compute the vector of associations $\mathbf{y}$ using a greedy approach, starting by matching the most probable fruits first and removing them from the pool of candidates.
This process continues until all fruits in $\mathcal{F}^t$ have been matched or labeled as unmatched.

\subsection{Loss Function}
We trained the fruit instance segmentation method following Marcuzzi~\etalcite{marcuzzi2023ral}, using a loss function $\mathcal{L}_{\text{ins}}$ composed of two terms: 
\begin{equation}
\mathcal{L}_{\text{ins}} = \mathcal{L}_{\text{sem}} + \lambda_{\text{off}} \mathcal{L}_{\text{off}}.
\end{equation}

The first term, $\mathcal{L}_{\text{sem}}$, accounts for the semantic segmentation of fruits and is a weighted sum of a cross-entropy loss~$\mathcal{L}_{\text{ce}}$ and a Lovász-Softmax loss~\cite{berman2018cvpr}~$\mathcal{L}_{\text{Lovász}}$:
\begin{equation}
\mathcal{L}_{\text{sem}} = \lambda_{\text{ce}} \mathcal{L}_{\text{ce}} + \lambda_{\text{Lov}} \mathcal{L}_{\text{Lovász}}.
\end{equation}

The second term, $\mathcal{L}_{\text{off}}$, measures the discrepancy between the predicted and ground truth offsets.
It is given by:
\begin{equation}
\mathcal{L}_{\text{off}} = \frac{1}{|\mathcal{P}_f|} \sum_{j=1}^{|\mathcal{P}_f|}{\| \mathbf{o}_j - \mathbf{\hat{o}}_j\|_1},
\end{equation}
where $\mathbf{o}_j \in \mathbb{R}^3$ is the $j$-th predicted offset and $\mathbf{\hat{o}}_j \in \mathbb{R}^3$ is the corresponding ground truth offset.

We train our descriptor extraction and re-identification method end-to-end by batch matching with a weighted loss function,~$\mathcal{L}_\text{m}$, composed of two terms:
\begin{equation}
\mathcal{L}_\text{m} = \mathcal{L}_{\text{ce}} + \lambda_{\text{inj}} \mathcal{L}_{\text{inj}}.
\end{equation}

The first,~$\mathcal{L}_{\text{ce}}$, computes the cross-entropy loss between the ground truth matrix~$\mathbf{\hat{H}}$ and the predicted matrix~$\mathbf{H^*}$:
\begin{equation}
\mathcal{L}_{\text{ce}} = \sum_{i=1}^{A}{ \sum_{j=1}^{B+1}{ -\mathbf{\hat{H}}_{ij} \log(\mathbf{H^*}_{ij}) } }.
\end{equation}

The second, $\mathcal{L}_{\text{inj}}$, forces the network to learn a bijective function. In other words, the network should map distinct fruits of $\mathcal{F}^t$ to distinct fruits of $\mathcal{F}^{t-1}$ and vice versa, avoiding assigning multiple fruits from a particular set to the same fruit in the other set.
The loss $\mathcal{L}_{\text{inj}}$ ignores the fruit without match and is weighted by $\lambda_{\text{inj}}$ in the final loss computation $L_{\text{m}}$.
In particular, we compute $\mathcal{L}_{\text{inj}}$ as follows:
\begin{equation}
\mathcal{L}_{\text{inj}} = \lambda_{\text{inj}} (\mathcal{L}_{\text{row}} + \mathcal{L}_{\text{col}}),
\end{equation}
where
\begin{equation}
  \mathcal{L}_{\text{row}} = \sum_{i=1}^{A}{ \left| {(\sum_{j=1}^{B} { \mathbf{H*}_{ij} })} - 1 \right| }
\end{equation}
and
\begin{equation}
  \mathcal{L}_{\text{col}} = \sum_{j=1}^{B}{ \left| {(\sum_{i=1}^{A} { \mathbf{H*}_{ij} })} - 1 \right| }.
\end{equation}

\section{Experimental Evaluation}
\label{sec:exp}
%
The main focus of this work is a novel method for fruit monitoring based on fruit instance segmentation and re-identification on point clouds recorded at different moments.

We present our experiments to show the capabilities of our method, focusing on high-precision point clouds of fruits such as strawberries and apples. We recorded the strawberry point clouds with a terrestrial laser scanner in a greenhouse, while apple point clouds were obtained from RGB images using photogrammetric reconstruction.
The results of our experiments support our claims:
(i)~our approach is able to identify fruits in point clouds using an instance segmentation method;
(ii)~it outperforms baseline approaches  on the instance segmentation and re-identification tasks using real-world data.

\subsection{Dataset} \label{sec:dataset}

To evaluate the performance of our instance segmentation method, we conduct experiments on two datasets. The dataset presented by Riccardi~\etalcite{riccardi2023icra}, hereafter referred to as the "strawberry dataset", which serves as the primary benchmark and is also used for the downstream re-identification task.
We include the PFuji-Size dataset~\cite{genemola2021dib} only for instance segmentation comparative analysis against baseline methods.
These datasets vary in complexity and content (the first with strawberries, the second with apples), allowing us to assess both the robustness and generalizability of the proposed approach.

\nnew{Notably, the colored point clouds in both datasets are largely free of occlusions. This is achieved through multi-view integration: in the strawberry dataset, point clouds are obtained by aligning multiple point clouds acquired from different perspectives, while in the PFuji-Size dataset, point clouds are obtained using multi-view stereo algorithms applied to RGB images acquired from different viewpoints.}
\new{For more details about the datasets acquisition, please refer to their original works (Gené-Mola~\etalcite{genemola2021dib}, Riccardi~\etalcite{riccardi2023icra}).}

For the re-identification experiments, we focus exclusively on the strawberry dataset, as it provides the necessary annotations required for this task, \ie, instance segmentation coherence in time.
This dataset consists of point clouds collected with a high-precision Faro Focus3D-X130 laser scanner in a commercial greenhouse, containing the same row of strawberries at 3 different points in time, each separated by approximately one week. Let us call them $\mathcal{P}^1$, $\mathcal{P}^2$, and $\mathcal{P}^3$.
Each point cloud is associated with a set of ground truth fruit annotations (respectively, $\hat{\mathcal{F}^1}$, $\hat{\mathcal{F}^2}$, and $\hat{\mathcal{F}^3}$, containing 616, 556 and 159 strawberries).
Then, the fruits in $\hat{\mathcal{F}^2}$ are associated with the fruit in $\hat{\mathcal{F}^1}$. Let $\mathbf{\hat{y}}_{2, 1}$ be the ground truth vector of associations.
Similarly, fruits in $\hat{\mathcal{F}^3}$ are associated with fruits in $\hat{\mathcal{F}^2}$ with the vector of associations~$\mathbf{\hat{y}}_{3, 2}$.
In~$\mathbf{\hat{y}}_{2, 1}$, 56 strawberries from~$\hat{\mathcal{F}^2}$ are not matched to those in~$\hat{\mathcal{F}^1}$.
In~$\mathbf{\hat{y}}_{3, 2}$, 5 strawberries from~$\hat{\mathcal{F}^3}$ are not matched to~$\hat{\mathcal{F}^2}$. See details in~\tabref{tab:riccardi_dataset}.

\begin{table}[t]
    \centering
    \resizebox{1.0\columnwidth}{!}{%
    \begin{tabular}{cccccc}
        \toprule
        \multirow{2}{*}{Timepoint} & 
         Point & \multirow{2}{*}{Annot.} & \multirow{2}{*}{\# Annot.} & Matched & \multirow{2}{*}{\# Unmatched} \\
          & 
         Cloud &  &  & With &  \\
        \midrule
        $t = 1$ & $\mathcal{P}^1$ & $\hat{\mathcal{F}}^1$ & 616 & -- & -- \\
        $t = 2$ & $\mathcal{P}^2$ & $\hat{\mathcal{F}}^2$ & 556 & $\hat{\mathcal{F}}^1$ & 56 \\
        $t = 3$ & $\mathcal{P}^3$ & $\hat{\mathcal{F}}^3$ & 159 & $\hat{\mathcal{F}}^2$ & 5 \\
        \bottomrule
    \end{tabular}
    }
    \caption{Overview of the strawberry dataset. Each point cloud $\mathcal{P}^t$ corresponds to a different acquisition time. Ground truth fruit annotations are denoted as $\hat{\mathcal{F}}^t$.}
    \label{tab:riccardi_dataset}
\end{table}

PFuji-Size~\cite{genemola2021dib} is a publicly available dataset designed for the detection and sizing of fruits in agricultural settings. It comprises high-resolution RGB images and \nnew{3D point clouds derived from photogrammetry of Fuji apple trees}, captured under field conditions. Each scene is annotated with ground truth fruit instance descriptions (center and radius). It is divided into two main acquisitions, the first captured in 2018 and the second captured in 2020, both featuring three Fuji trees. We used the 2018 acquisition for training and validation, and the 2020 acquisition for testing.
While manually inspecting the annotations of the dataset, we noticed that some fruits were missing from the first two trees of the 2020 collection. For this reason, we only used the third tree's fruit annotations for testing.

\subsection{Metrics} \label{sec:metrics}
To evaluate the instance segmentation capabilities we use as metrics panoptic quality (PQ)~\cite{kirillov2019cvpr-ps}, consisting of segmentation quality (SQ) and recognition quality (RQ), and intersection over union (IoU).
IoU is defined by
\begin{equation} \label{eq:iou}
\text{IoU} = \frac{1}{|TP|} \sum_{(p, g) \in TP} \frac{|p \cap g|}{|p \cup g|},
\end{equation}
where $p$ is the set of points belonging to a predicted instance, $g$ is the set of points belonging to a ground truth instance, and $TP$ is the set of pairs of predicted and ground truth instances matched (with $\frac{|p \cap g|}{|p \cup g|} \geq 0.5$ as commonly done).

The metrics PQ, SQ, and RQ are defined as
\begin{equation} \label{eq:pq}
\text{PQ} = \underbrace{\frac{\sum_{(p, g) \in TP_\text{seg}} \text{IoU}(p, g)}{|TP_\text{seg}|}}_{\text{segmentation quality (SQ)}} \cdot \underbrace{\frac{|TP_\text{seg}|}{|TP_\text{seg}| + \frac{1}{2}|FP_\text{seg}| + \frac{1}{2}|FN_\text{seg}|}}_{\text{recognition quality (RQ)}},
\end{equation}
where true positives~($TP_\text{seg}$), false positives~($FP_\text{seg}$), and false negatives~($FN_\text{seg}$) represent matched pairs of segments, unmatched predicted segments, and unmatched ground truth segments, respectively~\cite{kirillov2019cvpr}.

To evaluate matching performance, we first compute the following metrics:
correct matching ($\text{CM}$, correctly matching two strawberries),
mismatching ($\text{MM}$, correctly detecting a strawberry as a matching one but relating it with the wrong corresponding strawberry),
false matching ($\text{FM}$, incorrectly labeling a no-matching strawberry with a strawberry),
true negative ($\text{TN}$, correctly labeling a strawberry with the no-match label),
and false negative ($\text{FN}$, incorrectly labeling a matching strawberry with the no-match label).
Based on these, we calculate the following scores: 
\begin{equation}
\text{F1\textsubscript{p}} = \frac{2\text{CM}}{2\text{CM} + \text{MM} + \text{FM} + \text{FN}},
\end{equation}
\begin{equation}
\text{F1\textsubscript{n}} = \frac{2\text{TN}}{2\text{TN} + \text{FM} + \text{FN}},
\end{equation}
\begin{equation} \label{mf1}
\text{mF1} = \frac{\text{F1\textsubscript{p}} + \text{F1\textsubscript{n}}}{2},
\end{equation}

The standard $\text{F1}$ score measures the goodness of the predictive capacity of a model, but focuses only on the positive class. By averaging the~F1\textsubscript{p} and~F1\textsubscript{n} scores in~\eqref{mf1}, we get an indicator for both, $\text{mF1}$.

In our experiments, we trained the re-identification module to maximize the $\text{mF1}$ score in the validation set, in order to balance performance between correct matching and correct no-match predictions.


\subsection{Implementation Details} \label{sec:implementation-details}

We trained the two tasks, \ie, instance segmentation and re-identification, separately, and in particular, we trained for the re-identification task only using the strawberry ground truth fruit annotations. 
This procedure enables a clear separation between the two modules, allowing for easy integration and replacement of the segmentation component with alternative solutions. This flexibility makes our method highly adaptable and appealing for various applications.

We trained from scratch both instance segmentation methods, MinkPanoptic (ours) and Superpoint Transformer, and on both datasets.

In the strawberry dataset, we trained MinkPanoptic using an initial learning rate of~0.01 linearly decreased, at each epoch, with a decay coefficient of~0.97.
In the apple dataset, we used an initial learning rate of~0.03 linearly decreased, at each epoch, with a decay coefficient of~0.97.
In both datasets, we trained Superpoint Transformer using the default learning rate parameters, \ie, an initial value of~0.01 with a cosine annealing scheduler with warmup.

We used the same, standard data augmentation techniques for both methods, such as applying a random yaw rotation, X or Y axis flip, X or Y scale change (with a uniformly random scaling factor in the range~[0.97, 1.03] for strawberry and~[0.95, 1.05] for apples), and point jittering (adding per-point noise with a normal distribution centered at zero and with variance 0.03\,m for strawberry and~0.1\,m for apples). 

For MinkPanoptic, we optimized the bandwidth values using the validation set, finding the best value to be of~0.01125\,m for the strawberry dataset and~0.035\,m for the apple dataset (see~\secref{sec:bandwidth}).

Also our descriptor extraction and matching module is trained from scratch. In the loss function, we used as weights~\mbox{$\lambda_{\text{ce}}=2$}, \mbox{$\lambda_{\text{Lov}}=10$}, \mbox{$\lambda_{\text{off}}=10$} and \mbox{$\lambda_{\text{inj}}=0.08$}.
We set the support radius \mbox{$s=0.2$\,m} and the voxel size \mbox{$v=5\cdot10^{-4}$\,m} for the fruit descriptor extraction module and the maximum matching distance \mbox{$h=0.05$\,m} for the fruit descriptor matching module based on empirical evaluation. 
We used \mbox{$L=4$} layers using hidden dimensions 8, 8, 16, 16, and 64 (8 is the number of channels produced by the convolution of the initial feature extraction, while the subsequent are the produced number of channels for each layer). 
\nnew{We set the input channel dimension of} the matching module transformer layer~$l$ to 512, the feedforward dimension to 1024, and the number of heads to 8.

We trained the descriptor extraction and matching module with a fixed learning rate of~\mbox{$3\cdot10^{-4}$}.
We augmented the dataset by applying to each input fruit point cloud a random rotation in the range [-30\textdegree, 30\textdegree] on each of the three axes, a point jittering added per point drawn from a normal distribution centered at zero and with variance~\mbox{$7\cdot10^{-4}$\,m}, and a color jittering, modifying each RGB channel with a random gaussian noise drawn from a normal distribution centered at zero and with variance 0.05.

To train the fruit descriptor extraction and re-identification model, we manually divided the training set $\bigl( \hat{\mathcal{F}^1}$,~$\hat{\mathcal{F}^2}$, and~$\mathbf{\hat{y}}_{2, 1} \bigl)$ into two non-overlapping sets, grouping corresponding fruits based on their 3D position.
We used the first group as a training set, containing approximately~80\,\% of strawberries, and the second as a validation set.

\subsection{Instance Segmentation Results}
\begin{table}[t]
\centering
\resizebox{1.0\columnwidth}{!}{%
\begin{tabular}{ccc|cccc}
\toprule
\textbf{dataset} & \textbf{method} & \textbf{class} & \textbf{IoU} & \textbf{RQ} & \textbf{SQ} & \textbf{PQ}  \\
\midrule
\multirow{6}{*}{\rotatebox[origin=c]{90}{\textbf{strawberry}}}
 & \multirow{3}{*}{\minitab[c]{superpoint\\transformer}} & background & 97.1 & \textbf{100} & 97.1 & 97.1 \\
 &  & strawberry & 48.3 & 48.8 & 79.2 & 38.7 \\\text
 &  & average & 72.7 & 74.4 & 88.1 & 67.9 \\
\cmidrule(lr){2-7}
 & \multirow{3}{*}{\minitab[c]{MinkPanoptic\\(ours)}} & background & 98.9 & \textbf{100} & 98.9 & 98.9 \\
 &  & strawberry & 80.9 & 75.8  & 84.3 & 63.8 \\
 &  & average & 89.9 & 87.9  & 91.6 & 81.4 \\

\midrule

\multirow{6}{*}{\rotatebox[origin=c]{90}{\textbf{PFuji-Size}}} & \multirow{3}{*}{\minitab[c]{superpoint\\transformer}} & background & 92.5 & \textbf{100} & 92.2 & 92.2 \\
 &  & apple & 39.4 & 49.2 & 71.0 & 34.9 \\
 &  & average & 65.9 & 74.6 & 81.6 & 63.6 \\
\cmidrule(lr){2-7}
 & \multirow{3}{*}{\minitab[c]{MinkPanoptic\\(ours)}} & background & 94.6  &  \textbf{100} &  96.6  &  96.6 \\
 &  & apple & 50.1  &  52.1  &  78.3  &  40.7 \\
 &  & average & 72.3  &  76.0  &  87.4  &  68.7 \\
\midrule

\multirow{6}{*}{\rotatebox[origin=c]{90}{\textbf{average}}} & \multirow{3}{*}{\minitab[c]{superpoint\\transformer}} & background & 94.8 & \textbf{100} & 94.7 & 94.7 \\
 &  & fruit & 43.9  & 49  & 75.1  & 36.8 \\
 &  & average & 69.3  & 74.5  & 84.9  & 65.8 \\
\cmidrule(lr){2-7}
 & \multirow{3}{*}{\minitab[c]{MinkPanoptic\\(ours)}} & background & \textbf{96.8}  & \textbf{100} & \textbf{97.8}  & \textbf{97.8} \\
 &  & fruit & \textbf{65.5} & \textbf{64.0} & \textbf{81.3} & \textbf{52.3} \\
 &  & average & \textbf{81.1} & \textbf{82.0} & \textbf{89.5} & \textbf{75.1} \\
\bottomrule
\end{tabular}
}
\caption{Comparison of Superpoint Transformer~\cite{robert20243dv} and our MinkPanoptic on the instance segmentation task. All values are in~\%.}
\label{tab:instsegm_metrics}
\end{table}

We compared our instance segmentation method, MinkPanoptic, with multiple baselines and on two fruit point cloud datasets.
Surprisingly, \new{current} state-of-the-art instance segmentation methods struggle when trained with small or very small data, providing poor results in a seemingly simple task (see~\secref{sec:ins_sota}).
Only Superpoint Transformer~\cite{robert20243dv} was able to accurately segment strawberry or apple instances, and for this reason, we only report the comparison with this baseline.

In the strawberry dataset, we trained on the first two point clouds, \ie, $\mathcal{P}^1$ as the actual training set and $\mathcal{P}^2$ as the validation set, and tested on the third point cloud, \ie, $\mathcal{P}^3$.

Due to the pre-processing requirements of the Superpoint Transformer, we constructed a dataset by extracting~800 annotated point clouds for training and~200 for validation from the three original point clouds. Each sample was generated by cropping around a randomly selected seed point within the strawberry row, resulting in segments~0.15\,m wide, with a voxel size of~0.001\,m.
In contrast, MinkPanoptic was trained using the same number of point clouds, but with a larger crop width of~0.3\,m and a finer voxel size of~0.0005\,m, made possible by its lower memory usage, particularly during training.

\begin{figure}[p]
  \centering
  \includegraphics[width=1.0\linewidth]{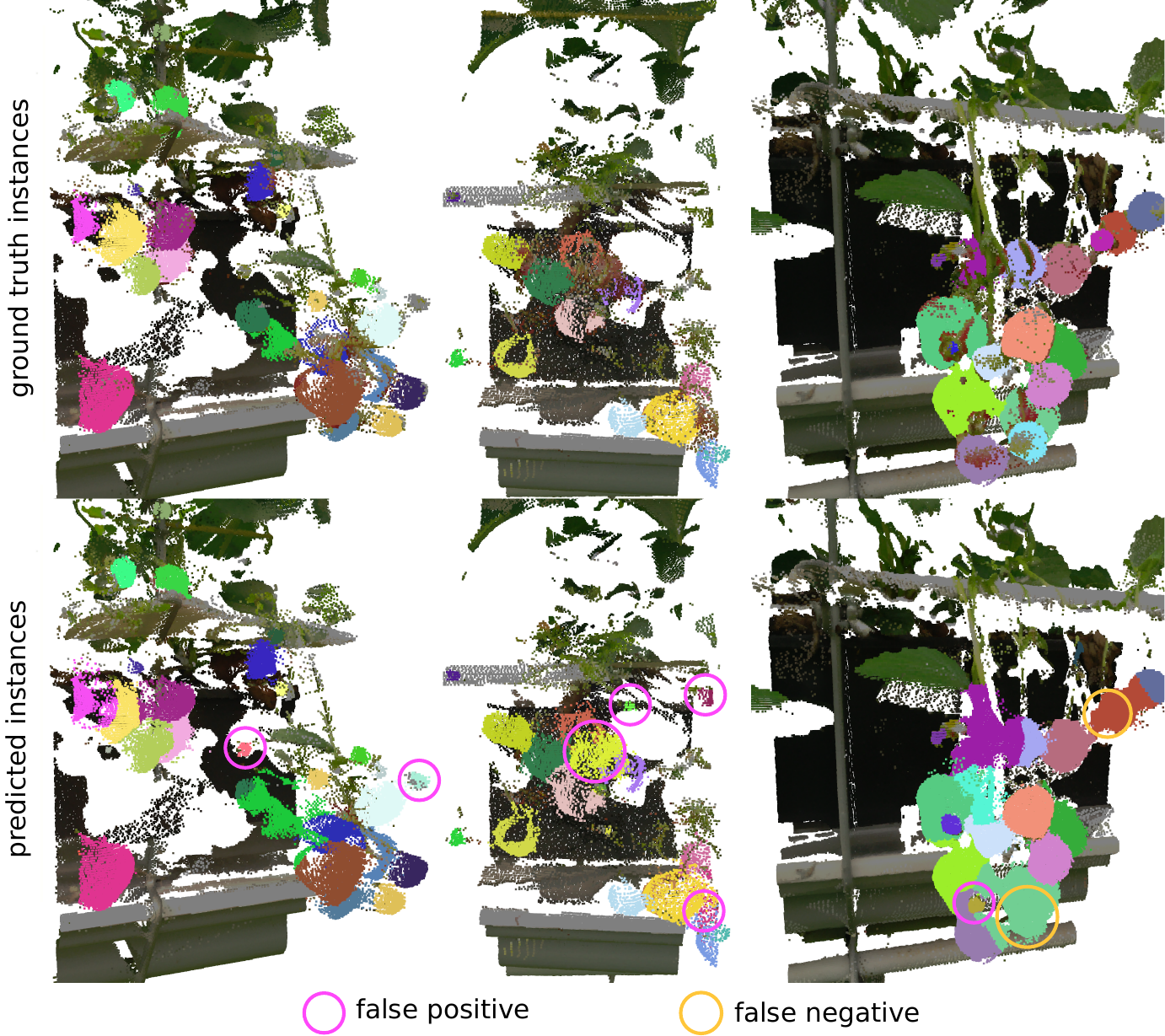}
  \caption{\nnew{Three qualitative examples of instance segmentation.
  On top is the ground truth, while below is the predicted segmentation.
  Identical instance color between the two rows indicates correct detections with at least 50\,\% IoU overlap.
  False positives (highlighted with \textcolor{circle_fuchsia}{fuchsia circles}) are fruits that were not present in the ground truth dataset (being too noisy to be clearly labeled) or leaves that were incorrectly classified as fruits, while false negatives (highlighted with \textcolor{circle_orange}{yellow circles}) are missed detections of fruits.
  On the left and center, two situations in which false positives are predominant. On the right, false negatives are predominant.}}
  \label{fig:segmentation_failures}
\end{figure}

\begin{figure}[p]
  \centering
  \includegraphics[width=1.0\linewidth]{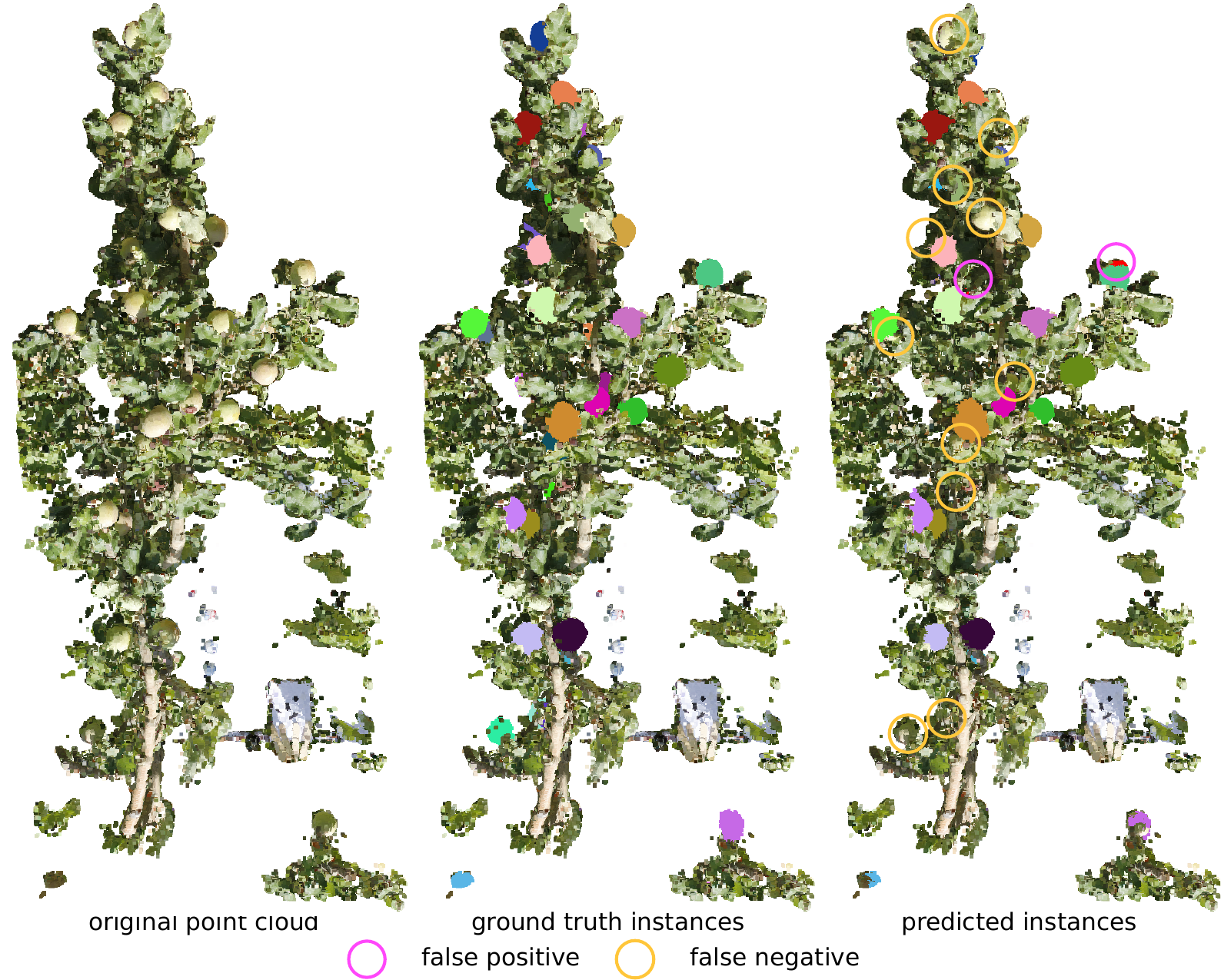}
  \caption{\nnew{Instance segmentation results on the PFuji-Size apple dataset.
  Although apparent occlusions are visible from this single viewpoint, the 3D point cloud is obtained via multi-view RGB fusion, which substantially reduces the single-view occlusions in the reconstructed data.
  Identical instance color between the two rows indicates correct detections with at least 50\,\% IoU overlap.
  False positives are highlighted with \textcolor{circle_fuchsia}{fuchsia circles} and false negatives with \textcolor{circle_orange}{yellow circles}.
  On the left is the original point cloud, on the center the ground truth instance segmentation, on the right the predicted segmentation.}}
  \label{fig:segmentation_failures_apples}
\end{figure}

The results are summarized in~\tabref{tab:instsegm_metrics}.
Across all datasets, MinkPanoptic consistently achieves the highest performance among the compared instance segmentation methods. 
The improvement is particularly evident in the strawberry dataset, especially for the strawberry class, where the gains are substantial. This may be attributed to the spherical nature of the fruit, which aligns well with MinkPanoptic's architectural strengths, specifically, its offset prediction mechanism and mean shift clustering, both of which are well-suited to delineating compact, rounded instances.
MinkPanoptic also \nnew{shows} superior performance in the apple dataset, further confirming its robustness across different fruit types.
In particular, all methods achieve a recognition quality of~100.0\,\% for the background class, indicating that all ground truth background points were correctly predicted as background points in these regions.
In general, substantial improvements in panoptic quality and intersection over union for the fruit class highlight how MinkPanoptic’s design and training strategy make it particularly effective for high-precision segmentation in agricultural scenarios.
\nnew{We show qualitative examples of the segmentation masks predicted by MinkPanoptic on the strawberry dataset in~\figref{fig:segmentation_failures} and on the Pfuji-Size dataset in~\figref{fig:segmentation_failures_apples}.}

\subsection{Discussion: Instance Segmentation state-of-the-art performance} \label{sec:ins_sota}
Surprisingly, \new{current} state-of-the-art instance segmentation methods (\ie,~\cite{xiao2025ijcv, schult2023icra, marcuzzi2023ral, kolodiazhnyi2024cvpr, shin2024cvpr}) struggle when trained with small or very small data, providing poor results in a seemingly simple task, \nnew{such as strawberry or apple segmentation}.
There are many reasons why.
Most methods are best suited for LiDAR point clouds, \eg P3Former~\cite{xiao2025ijcv}, exploiting their cylindrical space distribution.
\new{Methods like~Mask3D~\cite{schult2023icra}, MaskPLS~\cite{marcuzzi2023ral}, OneFormer3D~\cite{kolodiazhnyi2024cvpr}, and Spherical Mask~\cite{shin2024cvpr} suffer in this task due to their reliance on voxelization, which can lead to the loss of small object details during downsampling.
Using small voxels to preserve these details dramatically increases computational costs, making it impractical for our application, while increasing the voxel size to lower the computational complexity results in missing the small objects.
Moreover, the multi-resolution technique of these methods, usually helpful, deteriorates the performance on dense, small-scale datasets like strawberries, and the learned queries might collapse to similar representations, failing to differentiate between individual fruits.
We made every effort to include these state-of-the-art methods in our comparison; however, due to the severe limitations described above, their performance was so poor that they failed to produce even basic, usable segmentation results on our datasets.
}

\subsection{Re-identification Results}

In this experiment, we evaluate the performance of our approach in the re-identification task. It supports the claim that our method can identify fruits in point clouds using an instance segmentation method and outperforms baseline approaches on the re-identification task using real-world data collected in a greenhouse.
We will only use the strawberry dataset, as it provides the necessary annotations required for this task.
We match the fruit instance segmentation predicted on~$\mathcal{P}^3$ at the previous step with the ground fruit annotation of~$\mathcal{P}^2$. For comparison, we will also evaluate on ground truth fruit annotations.
To analyze robustness under varying IoU conditions, we evaluate matching performance at multiple thresholds ranging from 5\,\% to 30\,\%, with 5\,\% step, reflecting increasingly strict requirements for spatial consistency in 3D point clouds.
Higher IoU thresholds select only the predicted instances that are more similar to the ground truth annotation, introducing many negative instances (false positives from the instance segmentator), while lower IoU thresholds also associate low-quality instance prediction with ground truth annotation, reducing the number of negatives. 
Using more than 30\% IoU threshold would result in very few matching instances, making the evaluation less meaningful.
Each ground truth instance is assigned the ID of the predicted fruit instance with the highest IoU, provided that it exceeds the considered threshold. All not matching predicted fruit instances are assigned a new ID and thus labeled as negatives.

We consider as baseline methods a nearest neighbor approach fine-tuned with Optuna~\cite{akiba2019kddm} as well as Riccardi~\etalcite{riccardi2023icra}.
\new{We compare only with these two baselines because, to the best of our knowledge, Riccardi~\etal is the only publicly available work addressing fruit re-identification in point clouds, while the nearest neighbor approach is a simple yet effective baseline.
Also, Riccardi~\etal demonstrated superior capabilities compared to other traditional descriptor-based matchings.}

The first is purely based on the relative position between the sets of fruits and matches them by linking a fruit~$\mathcal{F}_i^t \in \mathcal{F}^t$ with the fruit~$\mathcal{F}_{*, NN}^{t-1} \in \mathcal{F}^{t-1}$ having the minimum Euclidean distance from~$\mathcal{F}_i^t$.
Formally:
\begin{equation}
\mathcal{F}_{*, NN}^{t-1} = \argmin_{j} \|\mathbf{c}_{i}^t - \mathbf{c}_j^{t-1}\|_2,
\end{equation}
where~$\mathbf{c}_i^t, \mathbf{c}_j^{t-1}$ are the center of the strawberries~$\mathcal{F}_i^t, \mathcal{F}_j^{t-1}$. 
The fruits in $\mathcal{F}^{t-1}$ can match only one fruit in $\mathcal{F}^t$, and vice versa.
This method can match fruits, but cannot detect the no-match case. For this reason, we used Optuna, an open-source optimization framework, to find the optimal threshold~$\epsilon \in \mathbb{R}$ for which two fruits at a distance greater than~$\epsilon$ should be considered unmatchable.
A fruit with no matchable counterpart is then considered unmatched.
In our experiment, we maximized the $\text{mF1}$ score in the training set and found the optimal value~$\epsilon^* = 0.033\,\text{m}$. 
Riccardi~\etalcite{riccardi2023icra} proposed a histogram descriptor based on the Euclidean distance between neighboring fruits. Considering a target fruit, they divide the 3D space around it into angular sectors and count how many fruits fall in each sector. In our experiments, we use the parameter setting suggested in the original implementation.

\begin{table*}[t]
  \centering
  
  \resizebox{1.0\linewidth}{!}{%
  \begin{tabular}{c|
                  ccccccccc|
                  ccccccccc}
    \toprule
    \multirow{3}{*}{\minitab[c]{\textbf{IoU}\\\textbf{level}}} &
    \multicolumn{9}{c|}{\textbf{using MinkPanoptic prediction}} &
    \multicolumn{9}{c}{\textbf{using SPT prediction}} \\
    \cmidrule(lr){2-10}
    \cmidrule(lr){11-19}
    & 
    \multicolumn{3}{c|}{\textbf{NN}} &
    \multicolumn{3}{c|}{\textbf{Riccardi~\etal}} &
    \multicolumn{3}{c|}{\textbf{Ours}} &
    \multicolumn{3}{c|}{\textbf{NN}} &
    \multicolumn{3}{c|}{\textbf{Riccardi~\etal}} &
    \multicolumn{3}{c}{\textbf{Ours}} \\
    \cmidrule(lr){2-4}
    \cmidrule(lr){5-7}
    \cmidrule(lr){8-10}
    \cmidrule(lr){11-13}
    \cmidrule(lr){14-16}
    \cmidrule(lr){17-19}
    & F1\textsubscript{p} & F1\textsubscript{n} & \textbf{mF1} &
      F1\textsubscript{p} & F1\textsubscript{n} & \textbf{mF1} &
      F1\textsubscript{p} & F1\textsubscript{n} & \textbf{mF1} &
      F1\textsubscript{p} & F1\textsubscript{n} & \textbf{mF1} &
      F1\textsubscript{p} & F1\textsubscript{n} & \textbf{mF1} &
      F1\textsubscript{p} & F1\textsubscript{n} & \textbf{mF1} \\
    \midrule
    \textbf{GT   }  & 84.7 & \textbf{75.0} & \textbf{79.9} & \textbf{92.6} & 0.0 & 46.3 & 76.4 & 51.4 & 63.9 & 84.7 & \textbf{75.0} & \textbf{79.9} & \textbf{92.6} & 0.0 & 46.3 & 76.4 & 51.4 & 63.9 \\
    \textbf{@5\% }  & 76.3 & 38.9 & 57.6 & 73.5 & 38.3 & 55.9 & \textbf{81.5} & \textbf{63.9} & \textbf{72.7} & \textbf{80.8} & 21.4 & 51.1 & 65.8 & 9.1 & 37.5 & 80.6 & \textbf{50.0} & \textbf{65.3} \\
    \textbf{@10\%}  & 74.9 & 35.4 & 55.1 & 71.5 & 35.6 & 53.6 & \textbf{80.0} & \textbf{59.6} & \textbf{69.8} & \textbf{78.0} & 15.4 & 46.7 & 69.9 & 6.1 & 38.0 & 76.2 & \textbf{39.2} & \textbf{57.7} \\
    \textbf{@15\%}  & 73.3 & 33.3 & 53.3 & 70.1 & 35.8 & 53.0 & \textbf{78.1} & \textbf{56.9} & \textbf{67.5} & \textbf{74.1} & 11.5 & 42.8 & 60.5 & 7.0 & 33.8 & 71.2 & \textbf{34.4} & \textbf{52.8} \\
    \textbf{@20\%}  & 71.8 & 30.4 & 51.1 & 68.8 & 36.8 & 52.8 & \textbf{75.7} & \textbf{53.0} & \textbf{64.4} & \textbf{67.2} & 8.7 & 38.0 & 55.0 & 8.1 & 31.6 & 64.9 & \textbf{27.2} & \textbf{46.0} \\
    \textbf{@25\%}  & 69.0 & 28.0 & 48.5 & 66.2 & 34.4 & 50.3 & \textbf{73.5} & \textbf{51.2} & \textbf{62.4} & \textbf{59.6} & 7.1 & 33.4 & 49.8 & 9.0 & 29.4 & 57.8 & \textbf{22.9} & \textbf{40.4} \\
    \textbf{@30\%}  & 61.1 & 22.8 & 41.9 & 60.6 & 34.5 & 47.5 & \textbf{65.7} & \textbf{44.6} & \textbf{55.2} & \textbf{51.6} & 5.9 & 28.8 & 38.1 & 7.5 & 22.8 & 50.3 & \textbf{21.2} & \textbf{35.7} \\
    \midrule
    \textbf{avg}    & 73.0 & 37.7 & 55.3 & 71.9 & 30.8 & 51.3 & \textbf{75.8} & \textbf{54.4} & \textbf{65.1} & \textbf{70.9} & 20.7 & 45.8 & 61.7 & 6.7 & 34.2 & 68.2 & \textbf{35.2} & \textbf{51.7} \\
    \bottomrule
  \end{tabular}
  }
  \caption{Re-identification performance across various IoU thresholds using different instance segmentation methods (MinkPanoptic~(ours) and Superpoint Transformer~\cite{robert20243dv}) and matching baselines (Nearest Neighbor~(NN), Riccardi~\etalcite{riccardi2023icra} and ours).}
  \label{tab:matching_all}
\end{table*}

\begin{figure*}[t]
  \centering
  \includegraphics[width=\linewidth]{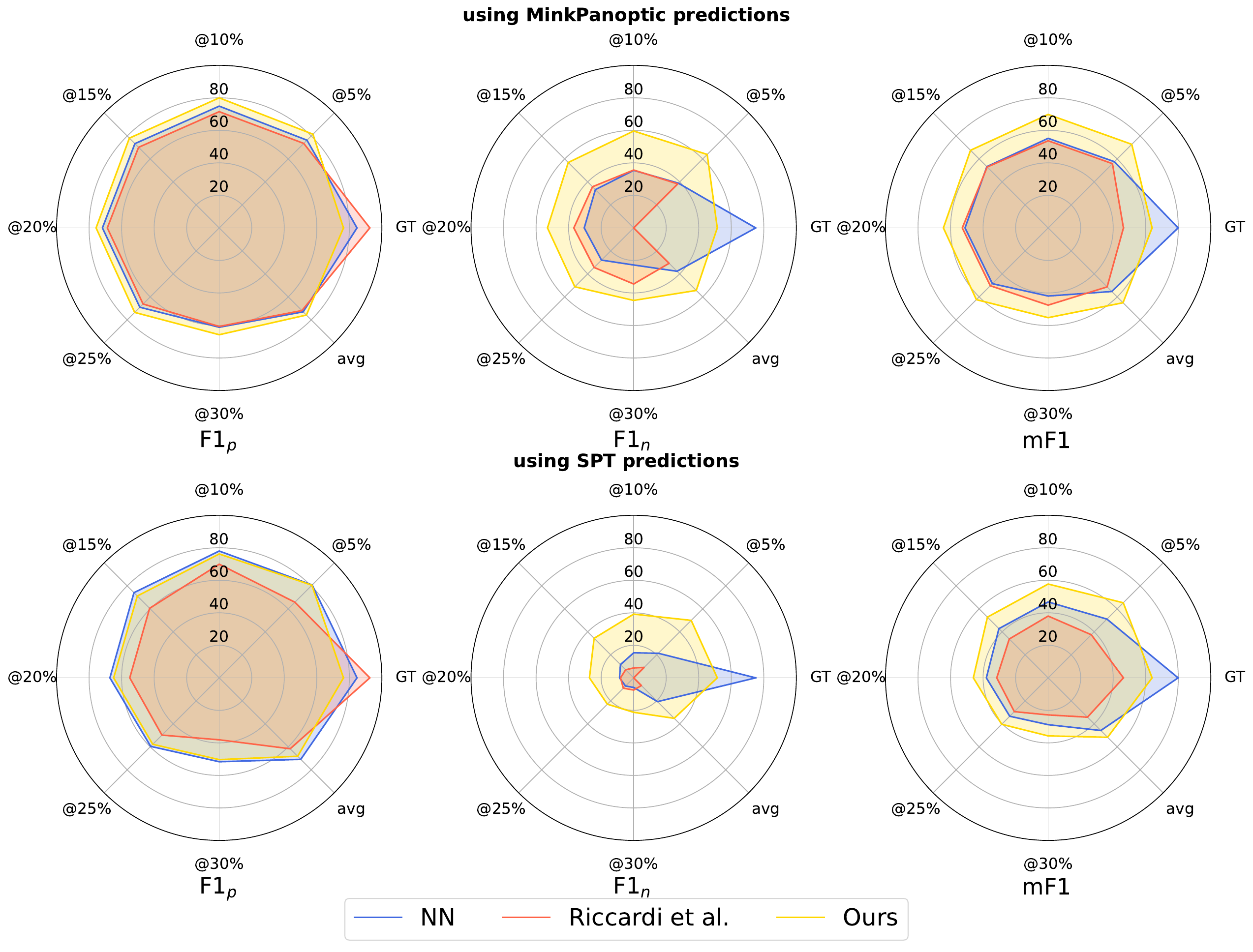}
  \caption{Radar plots comparing re-identification performance, detailed in~\tabref{tab:matching_all}, at varying IoU thresholds using instance segmentations from our method (MinkPanoptic) and SuperPoint. Best viewed in color.}
  \label{fig:radar_plot}
\end{figure*}

We compare our re-identification approach with baselines using both the instance segmentation prediction of MinkPanoptic and Superpoint Transformer. The results are reported in~\tabref{tab:matching_all} while in~\figref{fig:radar_plot} a visualization of the numerical results makes the comparison clearer.
With the instance segmentation provided by MinkPanoptic, our method consistently achieves the highest performance across all metrics in the test set using predicted instances, except when using ground truth annotations.
Our approach obtains an average score~$\text{mF1}$ of 65.1\,\%, outperforming the second-best method by 9.8\,\%.
The nearest neighbor approach generally surpasses Riccardi \etal.
With the instance segmentation provided by Superpoint Transformer, our method consistently achieves the highest performance across key metrics such as F1\textsubscript{n} and $\text{mF1}$. In particular, it attains an average $\text{mF1}$ score of 51.7\,\%, outperforming the second-best method by 5.9\,\%. While the nearest neighbor baseline achieves the highest F1\textsubscript{p}, it is generally more effective at correctly identifying positive matches, likely due to its reliance on spatial proximity, an advantage when instances remain near each other across time steps.
In contrast, our method excels in distinguishing negative matches, a critical advantage given the large number of negative pairs introduced by the relatively coarse instance segmentation from Superpoint Transformer. This balanced performance across both positive and negative samples makes our approach particularly robust, yielding the best overall results on this dataset.

\begin{figure*}[p]
  \centering
  \includegraphics[width=0.95\linewidth]{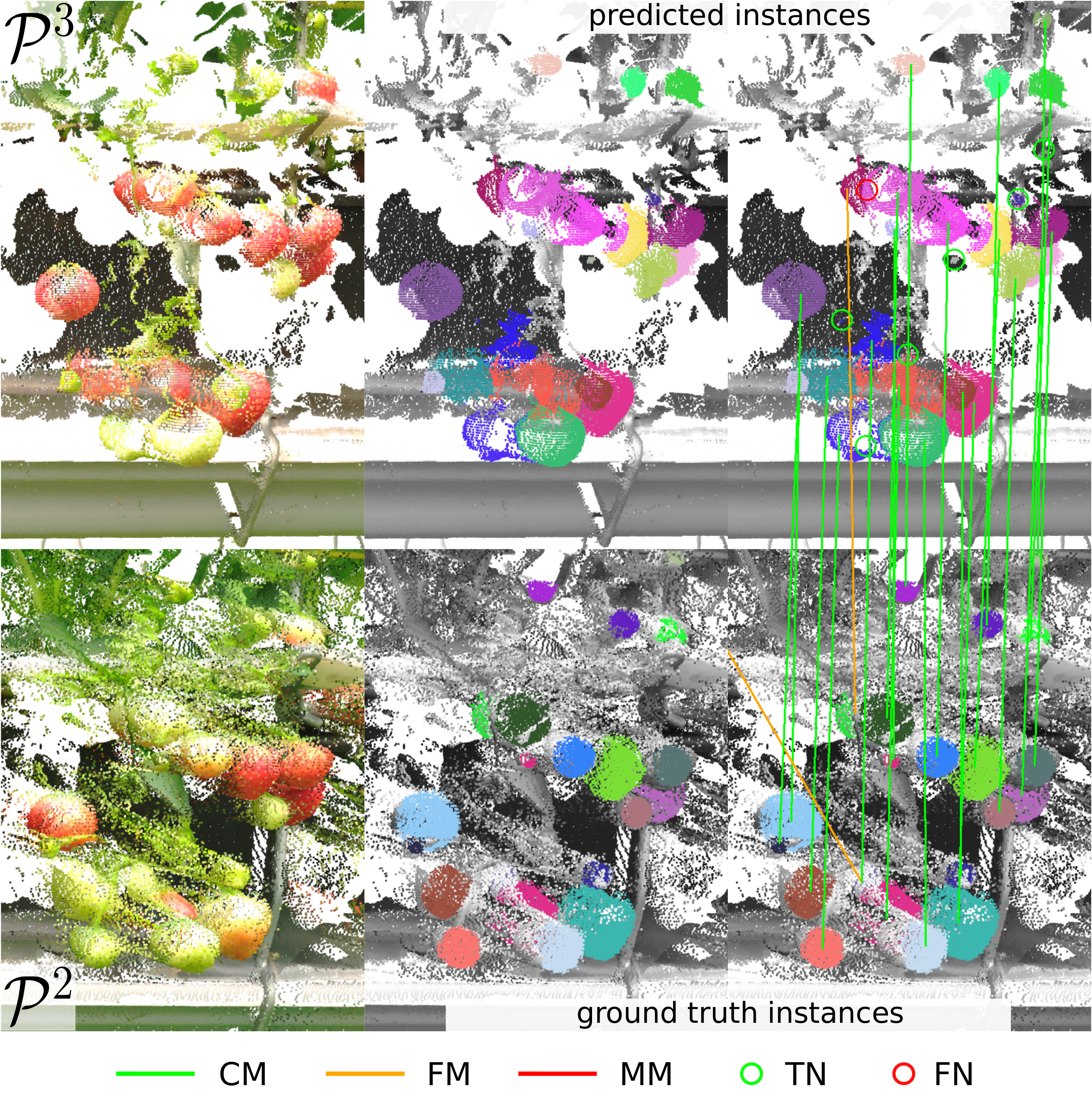}
  \caption{\new{Fruit instance segmentation and re-identification using our method, in a good performance scenario. 
    On the top are point clouds from~$\mathcal{P}^3$ that we match with the ground truth fruit annotations on~$\mathcal{P}^2$ depicted below.
    From left to right: the original, colored point clouds; the instance segmentation (ground truth for~$\mathcal{P}^2$, predicted using MinkPanoptic for~$\mathcal{P}^3$); the re-identification results.
    \textcolor{green}{Green lines} indicate correct matches~($\text{CMs}$), \textcolor{red}{red lines} indicate false matches~($\text{FMs}$), \textcolor{orange}{orange lines} indicate mismatches~($\text{MMs}$), \textcolor{green}{green circles} indicate true no-matches~($\text{TNs}$) and \textcolor{red}{red circles} indicate false no-matches~($\text{FNs}$).
    Most matches are correct, with only two~\textcolor{red}{$\text{FMs}$} and one~\textcolor{red}{$\text{FN}$}.
    Best viewed in color.}}
  \label{fig:instsegm_quality_good}
\end{figure*}

\begin{figure*}[p]
  \centering
  \includegraphics[width=0.95\linewidth]{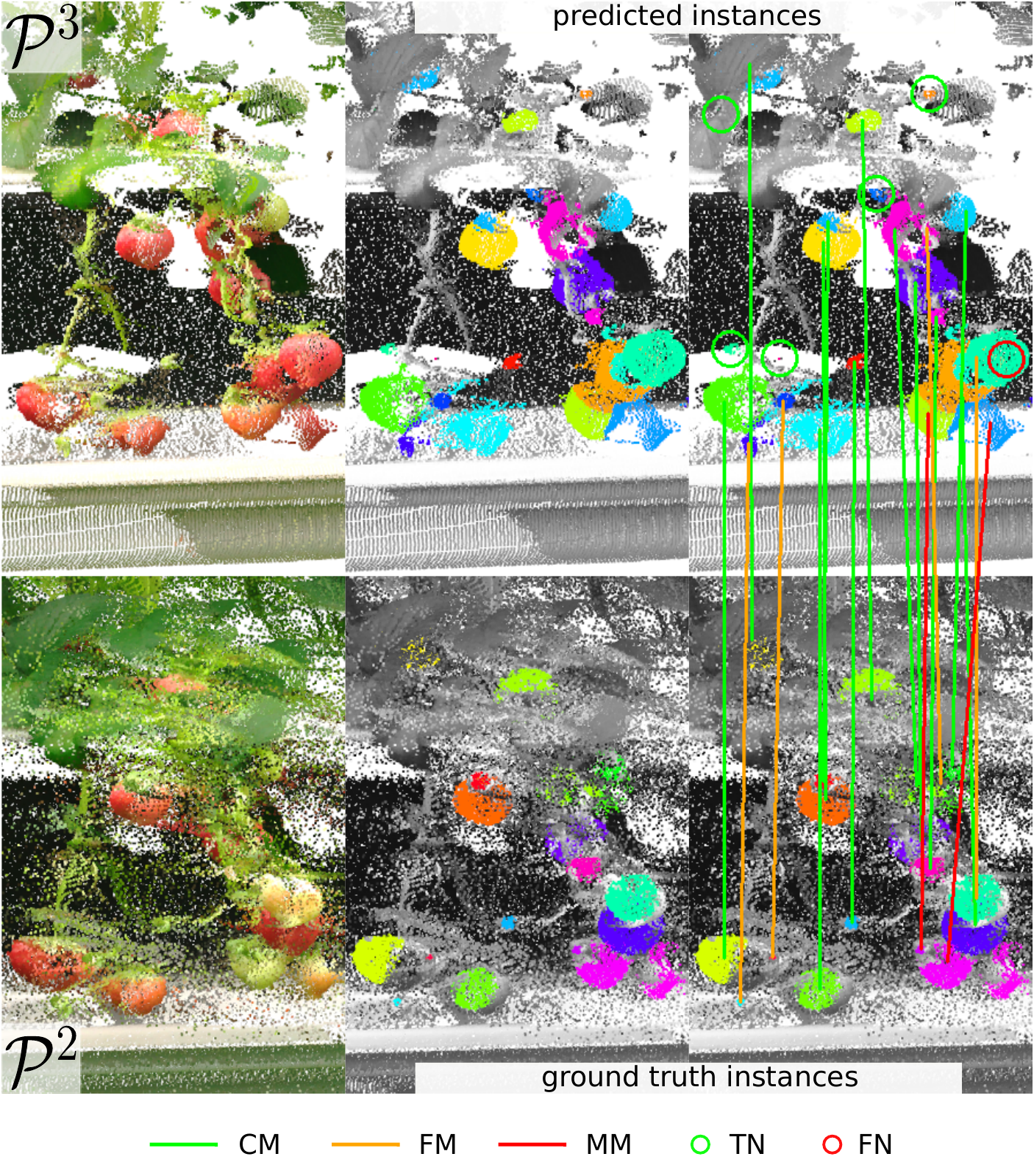}
  \caption{\new{Fruit instance segmentation and re-identification using our method, in a difficult scenario. 
  On the top are point clouds from~$\mathcal{P}^3$ that we match with the ground truth fruit annotations on~$\mathcal{P}^2$ depicted below.
  From left to right: the original, colored point clouds; the instance segmentation (ground truth for~$\mathcal{P}^2$, predicted using MinkPanoptic for~$\mathcal{P}^3$); the re-identification results.
  \textcolor{green}{Green lines} indicate correct matches~($\text{CMs}$), \textcolor{red}{red lines} indicate false matches~($\text{FMs}$), \textcolor{orange}{orange lines} indicate mismatches~($\text{MMs}$), \textcolor{green}{green circles} indicate true no-matches~($\text{TNs}$) and \textcolor{red}{red circles} indicate false no-matches~(*$\text{FNs}$).
  Many matches are correct, but there are also several~\textcolor{red}{$\text{FMs}$},~\textcolor{orange}{$\text{MMs}$} and one \textcolor{red}{$\text{FN}$}.
  Best viewed in color.}}
  \label{fig:instsegm_quality_lessgood}
\end{figure*}

\new{In \figref{fig:instsegm_quality_good} and \figref{fig:instsegm_quality_lessgood}, we show qualitative results of our re-identification method using the instance segmentation predicted by MinkPanoptic.
In \figref{fig:instsegm_quality_good}, the re-identification is almost perfect, with only two~$\text{FMs}$ and one~$\text{FN}$.
In \figref{fig:instsegm_quality_lessgood}, the re-identification is less accurate, with several~$\text{FMs}$, $\text{MMs}$, and one~$\text{FN}$. This is mainly due to the instance segmentation errors, such as merged fruits (leading to the~$\text{FNs}$ on bottom right) or false detections (leading to the~$\text{FNs}$ on bottom left).
Nonetheless, most detection faults are correctly handled by our re-identification method with several~$\text{TNs}$.
}
\figref{fig:radar_plot}~\new{quantitatively }highlights the superior performance of our approach. The radar plot of our method has always a greater area with respect to the baselines when using MinkPanoptic's predictions. With SPT's, the area is slightly smaller only on the F1\textsubscript{p} radar plot, while the area is sharply larger on the F1\textsubscript{n} and mF1 radar plots.

\new{For a broader discussion of the results, the method's limitations and future directions, refer to~\secref{sec:discussion}.}

%

\section{\nnew{Computational Efficiency and Runtime Analysis}}

\nnew{While the primary focus of this work is accuracy and robustness, computational efficiency is an important practical consideration for agricultural monitoring systems.
All experiments were conducted on a workstation equipped with an NVIDIA TITAN RTX\,GPU (24 GB memory) and an Intel Core i9-10920X CPU @ 3.50 GHz.}

\nnew{Training the instance segmentation module requires approximately 2 hours on average for both the strawberry and PFuji-Size apple datasets.
At inference time, the instance segmentation has a runtime of 100\,ms / 100k points.}

\nnew{Training the descriptor extraction and matching modules takes less than 1~hour.
At inference time, processing and matching 200 strawberries spanning approximately 2 m of crop row requires about 1.2\,s.
This runtime includes fruit support point cloud extraction and descriptor encoding of both point clouds (previous and current), and inference of the attention-based matching module.}

\nnew{Overall, the computational cost of the pipeline is dominated by the instance segmentation stage,
while the re-identification component scales approximately linearly with the number of detected fruits and remains lightweight in comparison.
The reported runtimes are well suited for offline and online phenotyping pipelines commonly adopted in agricultural monitoring and crop analysis applications.}

\section{Ablation Study}

\subsection{Instance Segmentation} \label{sec:bandwidth}

Our instance segmentation method, MinkPanoptic, is an end-to-end learned approach except for the fine-tuning of the bandwidth parameter of the mean shift algorithm, necessary for clustering the fruit instances.
We carefully investigated, using the validation set, the impact of different bandwidth values on performance.
\new{The bandwidth optimization process is a hyperparameter that is not involved in the training process of MinkPanoptic, since the loss only cares for binary semantic segmentation and offset prediction.
Thus, we were able to adjust it after the training by running multiple instance segmentations and analyzing the performance.
By doing so in a grid-search approach, we could identify the optimal bandwidth value that maximizes the~PQ on the validation set.}

\begin{table}[t]
  \centering
  \begin{tabular}{lccc|lccc}
    \toprule
    \multicolumn{4}{c|}{\textbf{strawberry}} & \multicolumn{4}{c}{\textbf{PFuji-Size}} \\
    \midrule
    \textbf{bw~(m)} & \textbf{RQ} & \textbf{SQ} & \textbf{PQ} & \textbf{bw~(m)} & \textbf{RQ} & \textbf{SQ} & \textbf{PQ} \\
    \midrule
    $\text{0.0094}^\dag$  & 80.2 & \textbf{86.2} & 71.5 & 0.02   & 80.2 & \textbf{85.3} & 70.4 \\
    0.0100  & 80.5 & \textbf{86.2} & 71.7 & 0.03   & 85.3 & 85.2 & 74.1 \\
    0.0105  & 80.8 & 86.1 & 71.9 & 0.0325 & 85.6 & 85.2 & 74.4 \\
    0.0110  & 80.8 & 86.1 & 71.9 & \textbf{0.035}  & 85.8 & 85.2 & \textbf{74.5} \\
    \textbf{0.01125} & \textbf{81.0} & 86.0 & \textbf{72.0} & 0.0375 & 85.7 & 85.1 & 74.3 \\
    0.0115  & 80.7 & 86.1 & 71.8 & 0.04   & 85.7 & 85.1 & 74.3 \\
    0.01175 & 80.8 & 86.1 & 71.9 & 0.0425 & \textbf{86.0} & 85.0 & 74.4 \\
    0.1200  & 80.8 & 86.1 & 71.9 & $\text{0.07}^\dag$  & 81.9 & 84.6 & 71.2 \\
    0.1300  & 80.3 & 86.1 & 71.5 & 0.09   & 74.6 & 84.5 & 65.7 \\
    \bottomrule
  \end{tabular}
  \caption{Effect of mean shift bandwidth fine-tuning on the instance segmentation performance on the strawberry and PFuji-Size datasets.
            Metrics are on the valid set in \%.
            \dag~symbol indicates the average radius size of the strawberries and apples in the valid set.}
  \label{tab:bandwidth_merged}
\end{table}

The results on the strawberry and PFuji-Size datasets are reported in~\tabref{tab:bandwidth_merged}.
The~PQ values followed a near-parabolic trend, allowing a clear selection of optimal values, which we used to evaluate our method, MinkPanoptic, in the test set.
Interestingly, for strawberry data, the bandwidth value 0.094\,m, which corresponds to the average radius size of the strawberries in the validation set, does not provide the best performance.
The same happens for the apple data. The 0.07\,m bandwidth, which corresponds to the average radius size of the apples in the validation set, does not provide the best performance.

\subsection{Re-identification}
\begin{figure*}[ht]
  \centering
  \includegraphics[width=\linewidth]{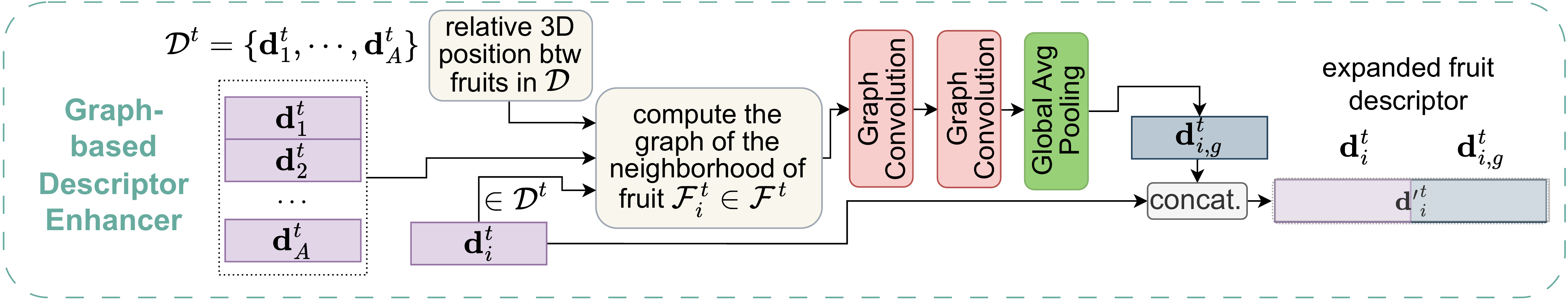}
  \caption{We investigated the significance of neighboring fruits in the descriptor computation. The module builds the graph of the neighborhood of fruit~$\mathcal{F}_i^t$. Two graph convolutions process the graph and a final global average pooling computes the neighborhood descriptor~$\mathbf{d}_{i,g}^t$, which is concatenated to the original~$\mathcal{F}_i$ fruit descriptor~$\mathbf{d}_i^t$.}
  \label{fig:graph-based-arch}
\end{figure*} 

\begin{figure}[ht]
  \centering
  \includegraphics[width=1.0\columnwidth]{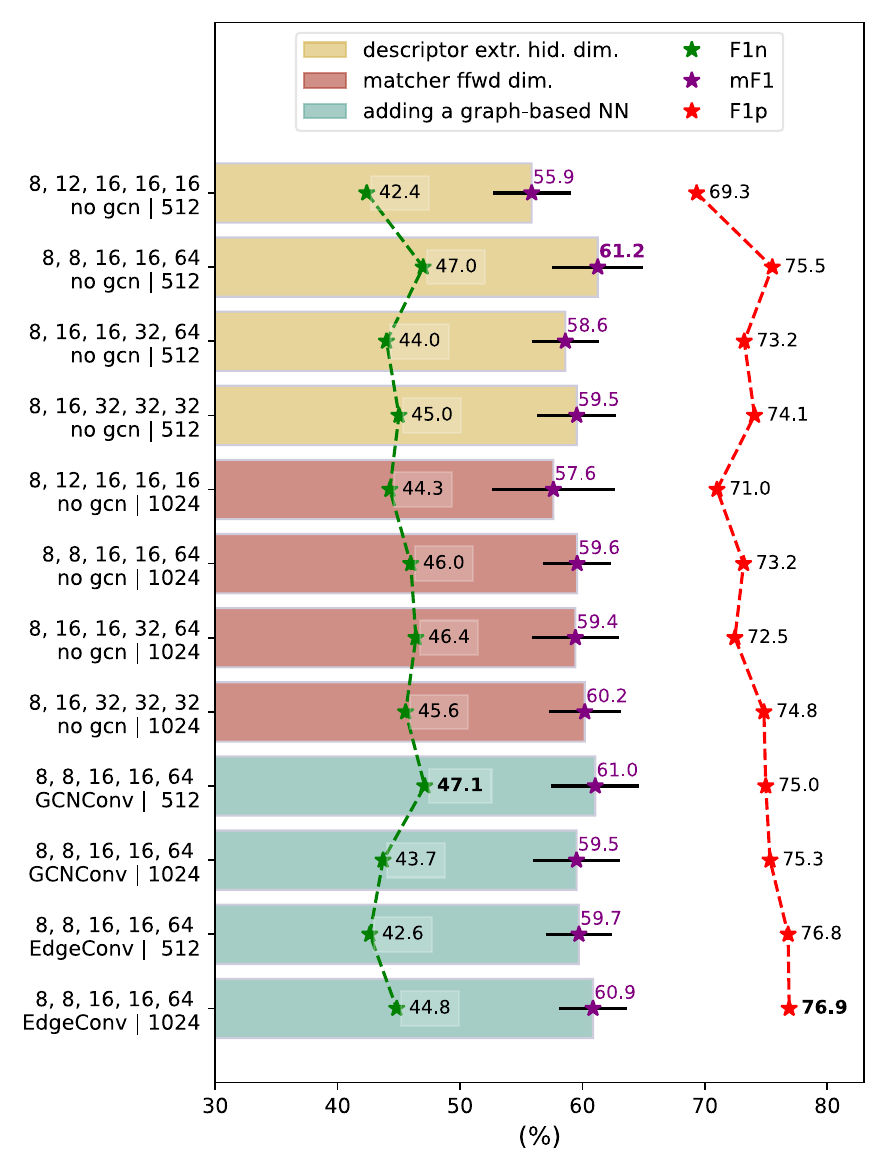}
  \caption{Ablation study to evaluate the impact of various design choices on our method.
      All values are in \%.
      The horizontal lines depict the standard deviation of $\text{mF1}$.
      \new{The first four bars colored in \textcolor{graph_yellow}{\rule[-0.07ex]{0.2cm}{1.6ex}} refer to the graph-free method with different fruit descriptor encoder hidden dimensions.
      The following four bars colored in \textcolor{graph_red}{\rule[-0.07ex]{0.2cm}{1.6ex}} refer to the graph-free method with different transformer feedforward dimensions.
      The last four bars colored in \textcolor{graph_green}{\rule[-0.07ex]{0.2cm}{1.6ex}} refer to the graph-based method with different graph convolutional operators and transformer feedforward dimensions.
      The best performing method for each of the F1\textsubscript{n}, F1\textsubscript{p} and~$\text{mF1}$ metrics is highlighted in bold.
      The~$\text{mF1}$ metric is the most important, as it balances positive and negative predictions, and is used to determine the overall best performing method.}
      }
  \label{fig:ablation_study}
\end{figure}

We conducted experiments to evaluate the impact of various \nnew{design choices of} fruit descriptor extractors on our method.
For each experiment, we performed a 5-fold cross-validation on the training set $\bigl(\hat{\mathcal{F}^1}$,~$\hat{\mathcal{F}^2}$, and~$\mathbf{\hat{y}}_{2, 1}\bigl)$ \nnew{which} we manually divided into five subsets based on the 3D position of strawberries. We tested different configurations of our fruit descriptor extraction and re-identification method, repeating each experiment four times with different seeds and averaging the results to minimize the effect of randomness.
The results are in~\figref{fig:ablation_study}.

Initially, we validated our re-identification method using a fixed matching module configuration, examining various hidden dimensions for the fruit encoder. 
The configuration with hidden dimensions [8, 8, 16, 16, 64] achieved the highest performance, with the best~$\text{mF1}$ score of 61.2\,\%, \mbox{F1\textsubscript{p} of 75.5\,\%} and~F1\textsubscript{n} of 47.0\,\%.

Next, we examined the effect of augmenting the transformer encoder layer's feedforward dimension from 512 to 1024 to determine if a more complex network could obtain better performance. The matcher with hidden dimensions [8, 16, 32, 32, 32] yielded the best in subset~$\text{mF1}$ score of 60.2\,\%, although this was still lower than the previous results.

Finally, we explore the importance of incorporating neighboring fruits in the descriptor computation.
We modified the descriptor extraction module to include a graph-based neural network that computes the neighboring fruits' descriptor and concatenates this to the original fruit descriptor. The architecture is depicted in~\figref{fig:graph-based-arch}.
We tested 2 graph convolutional operators: \mbox{GCNConv}~\cite{kipf2017iclr} and EdgeConv~\cite{wang2019tog}.
We kept fixed descriptor extractor hidden dimensions [8, 8, 16, 16, 64] and tried different matcher feedforward dimensions, \ie, 512 and 1024.

The method employing the GCNConv convolution operator, with feedforward dimension 512, obtained the best $\text{mF1}$ score of 61.0\,\% among methods using a graph convolutional operator (0.2\,\% less than the original, graph-free method) and the overall best F1\textsubscript{n} of 47.1\,\%.
In contrast, EdgeConv with the feedforward dimension 1024 obtains the overall best F1\textsubscript{p} of 76.9\,\%.
Although being best only in one metric, they do not balance the performance between negative and positive predictions, obtaining a lower $\text{mF1}$ score than the graph-free method.

In summary, our evaluation suggests that our graph-free method, composed of an encoder with hidden dimensions [8, 8, 16, 16, 64] and a matching module with a transformer's feedforward dimension 512, demonstrated to be the best performing in cross-validation, well balancing negative and positive predictions, and obtaining the best $\text{mF1}$.

\subsection{\new{Discussion of the Results, Limitations and Future Works}}
\label{sec:discussion}

\new{Our findings demonstrate the potential of leveraging high-resolution 3D point clouds for fruit monitoring, particularly in agricultural environments.
The proposed method effectively combines instance segmentation and re-identification, achieving state-of-the-art performance on challenging datasets.
However, several aspects need further discussion.
}

\new{First, the results highlight the importance of accurate instance segmentation as a precursor to successful re-identification.
While our method outperformed baselines, its performance is inherently tied to the quality of the instance segmentation.
Errors such as merged fruits or false detections can propagate to the re-identification stage, reducing overall accuracy.
This highlights the need for robust segmentation methods tailored to agricultural datasets, which often feature dense, occluded, and visually similar objects, unlike indoor or urban datasets.
}

\new{Second, the experiments reveal the adaptability of our approach to different fruit types and datasets.
The use of MinkPanoptic for instance segmentation proved effective across both strawberry and apple datasets, suggesting its generalizability.
However, the re-identification module was only evaluated on the strawberry dataset due to the lack of publicly available temporally consistent instance-annotated datasets for other fruits.
Expanding the evaluation to diverse datasets and fruit types would provide a more comprehensive understanding of the method's robustness and limitations.
}

\new{Third, the ablation study highlights the significance of design choices in the descriptor extraction and matching modules.
While incorporating graph-based features showed potential, the simpler graph-free approach achieved the best balance between positive and negative predictions.
This suggests that the added complexity of graph-based methods may not always translate to improved performance, particularly in scenarios with limited training data.
}

\new{Finally, the broader implications of this work extend beyond fruit monitoring.
The proposed method can be adapted to other domains requiring instance segmentation and temporal matching in 3D point clouds, such as forestry, construction, and urban planning.
However, the reliance on high-resolution LiDAR data may limit its applicability in resource-constrained settings.
Future work could explore the use of lower-cost sensors or hybrid approaches combining 2D and 3D data to enhance accessibility.
}

\section{Conclusion}
\label{sec:conclusion}
In this article, we presented a novel approach for temporal object instance analysis based on fruit instance segmentation and re-identification on colored point clouds acquired by a high-resolution LiDAR scanner at different points in time.
Our method first segments fruits using a learning-based instance segmentation approach, and then each segmented fruit is processed by a 3D sparse convolutional neural network to compute a compact descriptor.
Each fruit is matched with its corresponding instance from a previous data collection using its descriptors and an attention-based matching network designed for robust temporal association.
We implemented and evaluated our approach on real-world 3D datasets and provided comparisons with other existing techniques.
The experiments supported all claims made in this article and demonstrated that our method achieves superior performance in both instance segmentation and re-identification, even when trained on ground truth annotations but tested on predicted instance segmentation.
This flexibility makes our method highly adaptable to different scenarios, highlighting its potential for broader applications in 3D pattern recognition tasks where segmentation and temporal consistency are critical, allowing for easy integration and replacement of the segmentation component with alternative solutions.

\bibliographystyle{IEEEtran}
\bibliography{glorified, new}

@STRING{arpb                   = {Annual Review of Plant Biology}}

@STRING{arxiv                  = {arXiv preprint} }

@STRING{biosyseng              = {Biosystems Engineering} }

@STRING{case                   = {Proc.~of the Intl.~Conf.~on Automation Science and Engineering (CASE)} }

@STRING{cea                    = {Computers and Electronics in Agriculture} }

@STRING{cvpr                   = {Proc.~of the IEEE/CVF Conf.~on Computer Vision and Pattern Recognition (CVPR)} }

@STRING{cvprold                = {Proc.~of the IEEE Conf.~on Computer Vision and Pattern Recognition (CVPR)} }

@STRING{eccv                   = {Proc.~of the Europ.~Conf.~on Computer Vision (ECCV)} }

@STRING{iccvold                = {Proc.~of the IEEE Intl.~Conf.~on Computer Vision (ICCV)} }

@STRING{icml                   = {Proc.~of the Intl.~Conf.~on Machine Learning (ICML)} }

@STRING{icra                   = {Proc.~of the IEEE Intl.~Conf.~on Robotics \& Automation (ICRA)} }

@STRING{ifac                   = {IFAC-PapersOnLine}}

@STRING{ijcv                   = {Intl.~Journal~of Computer Vision (IJCV)} }

@STRING{iros                   = {Proc.~of the IEEE/RSJ Intl.~Conf.~on Intelligent Robots and Systems (IROS)} }

@STRING{jprs                   = {ISPRS Journal of Photogrammetry and Remote Sensing (JPRS)} }

@STRING{nipsjournel            = {Advances in Neural Information Processing Systems} }

@STRING{pami                   = {IEEE Trans.~on Pattern Analysis and Machine Intelligence (TPAMI)} }

@STRING{plosone                = {PLOS ONE} }

@STRING{ral                    = {IEEE Robotics and Automation Letters (RA-L)} }

@STRING{springer               = {Springer Verlag} }

@STRING{threedv                = {Proc.~of the Intl.~Conf.~on 3D Vision (3DV)} }

@STRING{tog                    = {ACM Trans.~on Graphics (TOG)} }

@STRING{tpami                  = {IEEE Trans.~on Pattern Analysis and Machine Intelligence (TPAMI)} }

@article{halstead2018ral,
  author  = {Halstead, Michael and McCool, Christopher and
             Denman, Simon and Perez, Tristan and Fookes, Clinton},
  journal = ral,
  number  = {4},
  pages   = {2995--3002},
  title   = {Fruit quantity and ripeness estimation using a
             robotic vision system},
  volume  = {3},
  year    = {2018},
}

@inproceedings{smitt2021icra,
  author    = {Smitt, Claus and Halstead, Michael and
               Zaenker, Tobias and Bennewitz, Maren and
               McCool, Chris},
  booktitle = icra,
  title     = {{PATHoBot}: {A} Robot for Glasshouse Crop Phenotyping and Intervention},
  year      = {2021},
  pages={2324--2330},
}

@inbook{stachniss2016handbook-slamchapter,
  author    = {C. Stachniss and J. Leonard and S. Thrun},
  title     = {{Springer Handbook of Robotics, 2nd edition}},
  chapter   = {Chapt.~46: Simultaneous Localization and Mapping},
  publisher = springer,
  year      = 2016,
  pages = {1153-1176}
}

@inproceedings{nuske2011iros,
  author    = {S. Nuske and S. Achar and T. Bates and S. Narasimhan and S. Singh},
  title     = {{Yield Estimation in Vineyards by Visual Grape Detection}},
  booktitle = iros,
  year      = 2011,
  pages={2352--2358},
}

@article{blok2021biosyseng,
  author  = {Blok, Pieter M and van Henten, Eldert J and
             van Evert, Frits K and Kootstra, Gert},
  journal = biosyseng,
  pages   = {213--233},
  title   = {Image-based size estimation of broccoli heads under
             varying degrees of occlusion},
  volume  = {208},
  year    = {2021}
}

@inproceedings{akiba2019kddm,
  title     = {Optuna: A Next-generation Hyperparameter Optimization Framework},
  author    = {Akiba, Takuya and Sano, Shotaro and Yanase, Toshihiko and Ohta, Takeru and Koyama, Masanori},
  booktitle = {Proc.~of the Intl.~Conf.~on Knowledge Discovery and Data Mining},
  year      = {2019},
  pages={2623--2631},
}

@inproceedings{berman2018cvpr,
  author    = {M. Berman and A. R. Triki and M. B. Blaschko},
  title     = {The lov{\'a}sz-softmax loss: A tractable surrogate for the optimization of the intersection-over-union measure in neural networks},
  booktitle = cvpr,
  year      = 2018,
  pages={4413--4421},
}

@inproceedings{graham2018cvpr,
  title     = {{3D Semantic Segmentation with Submanifold Sparse Convolutional Networks}},
  author    = {B. Graham and M. Engelcke and L. van der Maaten},
  booktitle = cvprold,
  pages={9224--9232},
  year      = {2018},
}

@article{vysotska2019ral,
  author  = {O. Vysotska and C. Stachniss},
  title   = {{Effective Visual Place Recognition Using Multi-Sequence Maps}},
  journal = ral,
  volume  = {4},
  number  = {2},
  pages   = {1730--1736},
  year    = 2019,
}

@inproceedings{kirillov2019cvpr,
  author    = {A. Kirillov and R. Girshick and K. He and P. Dollar},
  title     = {{Panoptic Feature Pyramid Networks}},
  booktitle = cvpr,
  year      = 2019,
  pages={6399--6408},
}

@inproceedings{kirillov2019cvpr-ps,
  author    = {A. Kirillov and K. He and R. Girshick and C. Rother and P. Doll{\'a}r},
  title     = {{Panoptic Segmentation}},
  booktitle = cvpr,
  year      = 2019,
  pages={9404--9413},
}

@inproceedings{he2017iccv-mr,
  author    = {K. He and G. Gkioxari and P. Doll{\'a}r and R. Girshick},
  title     = {{Mask R-CNN}},
  booktitle = iccvold,
  year      = 2017,
  pages={2961--2969},
}

@article{fiorani2013arpb,
  title    = {Future scenarios for plant phenotyping},
  author   = {F. Fiorani and U. Schurr},
  journal  = arpb,
  volume   = {64},
  pages    = {267--291},
  year     = {2013}
}

@inproceedings{mildenhall2020eccv,
  title     = {{NeRF: Representing Scenes as Neural Radiance Fields for View Synthesis}},
  author    = {B. Mildenhall and P.P. Srinivasan and M. Tancik and J.T. Barron and R. Ramamoorthi and R. Ng},
  booktitle = eccv,
  year      = {2020},
  pages={99--106},
}

@article{walter2018nas,
  author  = {A. Walter and R. Finger and R. Huber and N. Buchmann},
  title   = {Opinion: Smart farming is key to developing sustainable agriculture},
  volume  = {114},
  number  = {24},
  pages   = {6148--6150},
  year    = {2017},
  journal = {Proceedings of the National Academy of Sciences}
}

@article{duckett2018arxiv,
  author   = {T. Duckett and S. Pearson and S. Blackmore and B. Grieve and W. Chen and G. Cielniak and J. Cleaversmith and J. Dai and S. Davis and C. Fox and P. From and I. Georgilas and R. Gill and I. Gould and M. Hanheide and A. Hunter and F. Iida and L. Mihalyova and S. Nefti-Meziani and G. Neumann and P. Paoletti and T. Pridmore and D. Ross and M. Smith and M. Stoelen and M. Swainson and S. Wane and P. Wilson and I. Wright and G. Yang},
  title    = {{Agricultural Robotics: The Future of Robotic Agriculture}},
  journal  = arxiv,
  volume   = {arXiv:1806.06762},
  year     = 2018
}

@article{chebrolu2021plosone,
  author  = {N. Chebrolu and F. Magistri and T. L{\"a}be and C. Stachniss},
  title   = {{Registration of Spatio-Temporal Point Clouds of Plants for Phenotyping}},
  journal = plosone,
  year    = 2021,
  number  = {2},
  volume  = {16},
}

@inproceedings{ioffe2015icml,
  author    = {S. Ioffe and C. Szegedy},
  title     = {{Batch Normalization: Accelerating Deep Network Training by Reducing Internal Covariate Shift}},
  booktitle = icml,
  year      = {2015},
  pages={448--456},
}

@inproceedings{choy2019cvpr,
  author    = {C. Choy and J. Gwak and S. Savarese},
  title     = {{4D Spatio-Temporal ConvNets: Minkowski Convolutional Neural Networks}},
  booktitle = cvpr,
  year      = 2019,
  pages={3075--3084},
}

@article{comaniciu2002pami,
  title   = {{Mean Shift: A Robust Approach Toward Feature Space Analysis}},
  author  = {D. Comaniciu and P. Meer},
  journal = pami,
  year    = 2002,
  volume  = {24},
  number  = {5},
  pages   = {603--619},
}

@article{vaswani2017neurips,
  title     = {{Attention Is All You Need}},
  author={Vaswani, Ashish and Shazeer, Noam and Parmar, Niki and Uszkoreit, Jakob and Jones, Llion and Gomez, Aidan N and Kaiser, {\L}ukasz and Polosukhin, Illia},
  journal = nipsjournel,
  year      = 2017,
}

@inproceedings{riccardi2023icra,
  author    = {A. Riccardi and S. Kelly and E. Marks and F. Magistri and T. Guadagnino and J. Behley and M. Bennewitz and C. Stachniss},
  title     = {{Fruit Tracking Over Time Using High-Precision Point Clouds}},
  booktitle = icra,
  year      = {2023},
  pages={9630--9636},
}

@article{marcuzzi2023ral,
  author  = {R. Marcuzzi and L. Nunes and L. Wiesmann and J. Behley and C. Stachniss},
  title   = {{Mask-Based Panoptic LiDAR Segmentation for Autonomous Driving}},
  journal = ral,
  volume  = {8},
  number  = {2},
  pages   = {1141--1148},
  year    = 2023,
}

@article{wang2019tog,
author = {Wang, Yue and Sun, Yongbin and Liu, Ziwei and Sarma, Sanjay E. and Bronstein, Michael M. and Solomon, Justin M.},
title = {Dynamic Graph {CNN} for Learning on Point Clouds},
year = {2019},
volume = {38},
number = {5},
issn = {0730-0301},
doi = {10.1145/3326362},
journal = tog,
}

@article{kipf2017iclr,
  title={Semi-supervised Classification with Graph Convolutional Networks},
  author={Kipf, TN},
  journal={arXiv preprint arXiv:1609.02907},
  year={2016}
}

@article{watt2020arpb,
      author       = {Watt, Michelle and Fiorani, Fabio and Usadel, Björn and
                      Rascher, Uwe and Muller, Onno and Schurr, Ulrich},
      title        = {Phenotyping: New Windows into the Plant for Breeders},
      journal      = arpb,
      volume       = {71},
      number       = {1},
      issn         = {1545-2123},
      pages        = {},
      year         = {2020},
      doi          = {10.1146/annurev-arplant-042916-041124},
}

@article{weikuan2021cea,
title = {FoveaMask: A fast and accurate deep learning model for green fruit instance segmentation},
journal = cea,
volume = {191},
pages = {106488},
year = {2021},
issn = {0168-1699},
doi = {10.1016/j.compag.2021.106488},
author = {Weikuan Jia and Zhonghua Zhang and Wenjing Shao and Sujuan Hou and Ze Ji and Guoliang Liu and Xiang Yin},
}

@article{perez-borrero2020cea,
title = {A fast and accurate deep learning method for strawberry instance segmentation},
journal = cea,
volume = {178},
pages = {105736},
year = {2020},
issn = {0168-1699},
doi = {10.1016/j.compag.2020.105736},
author = {Isaac Pérez-Borrero and Diego Marín-Santos and Manuel E. Gegúndez-Arias and Estefanía Cortés-Ancos},
}

@article{kang2022cea,
title = {Accurate fruit localisation using high resolution LiDAR-camera fusion and instance segmentation},
journal = cea,
volume = {203},
pages = {107450},
year = {2022},
issn = {0168-1699},
doi = {10.1016/j.compag.2022.107450},
author = {Hanwen Kang and Xing Wang and Chao Chen},
}

@article{magistri2024ral,
author = {F. Magistri and Y. Pan and J. Bartels and J. Behley and C. Stachniss and C. Lehnert},
title = {{Improving Robotic Fruit Harvesting Within Cluttered Environments
Through 3D Shape Completion}},
journal = ral,
volume = {9},
number = {8},
pages = {7357-7364},
year = 2024,
doi = {10.1109/LRA.2024.3421788},
}

@article{ganesh2019ifac,
title = {Deep Orange: Mask {R-CNN} based Orange Detection and Segmentation},
author = {P. Ganesh and K. Volle and T.F. Burks and S.S. Mehta},
journal = {IFAC Proceedings Volumes},
volume = {52},
number = {30},
pages = {70-75},
year = {2019},
doi = {10.1016/j.ifacol.2019.12.499},
}

@article{gonzalez2019ieeeaccess,
  author={Gonzalez, Sebastian and Arellano, Claudia and Tapia, Juan E.},
  journal={IEEE Access}, 
  title={Deepblueberry: Quantification of Blueberries in the Wild Using Instance Segmentation}, 
  year={2019},
  volume={7},
  doi={10.1109/ACCESS.2019.2933062}
  }

@article{genemola2020cea,
title = {Fruit detection and 3D location using instance segmentation neural networks and structure-from-motion photogrammetry},
author = {Jordi Gené-Mola and Ricardo Sanz-Cortiella and Joan R. Rosell-Polo and Josep-Ramon Morros and Javier Ruiz-Hidalgo and Verónica Vilaplana and Eduard Gregorio},
journal = cea,
volume = {169},
pages = {105165},
year = {2020},
doi = {10.1016/j.compag.2019.105165},
}

@article{zhu2022tpami,
  author={Zhu, Xinge and Zhou, Hui and Wang, Tai and Hong, Fangzhou and Li, Wei and Ma, Yuexin and Li, Hongsheng and Yang, Ruigang and Lin, Dahua},
  journal=tpami, 
  title={Cylindrical and Asymmetrical 3D Convolution Networks for LiDAR-Based Perception}, 
  year={2022},
  volume={44},
  number={10},
  pages={6807-6822},
  doi={10.1109/TPAMI.2021.3098789}
  }

@article{kierdorf2021fai,
  author={Kierdorf, Jana  and Weber, Immanuel  and Kicherer, Anna  and Zabawa, Laura  and Drees, Lukas  and Roscher, Ribana },
  title={Behind the Leaves: Estimation of Occluded Grapevine Berries With Conditional Generative Adversarial Networks},
  journal={Frontiers in Artificial Intelligence},
  volume={5},
  year={2022},
  doi={10.3389/frai.2022.830026},
  issn={2624-8212},
}

@article{gomez2021ral,
  author={Gomez, Adrian Salazar and Aptoula, Erchan and Parsons, Simon and Bosilj, Petra},
  journal=ral, 
  title={Deep Regression Versus Detection for Counting in Robotic Phenotyping}, 
  year={2021},
  volume={6},
  number={2},
  pages={2902-2907},
  doi={10.1109/LRA.2021.3062586}
  }

@article{hao2024be,
title = {Automatic acquisition, analysis and wilting measurement of cotton 3D phenotype based on point cloud},
author = {Haoyuan Hao and Sheng Wu and YuanKun Li and Weiliang Wen and jiangchuan Fan and Yongjiang Zhang and Lvhan Zhuang and Longqin Xu and Hongxin Li and Xinyu Guo and Shuangyin Liu},
journal = biosyseng,
volume = {239},
pages = {173-189},
year = {2024},
issn = {1537-5110},
doi = {10.1016/j.biosystemseng.2024.02.010},
}

@article{boogaard2023be,
  title = {The added value of 3D point clouds for digital plant phenotyping – A case study on internode length measurements in cucumber},
  journal = {Biosystems Engineering},
  volume = {234},
  pages = {1-12},
  year = {2023},
  issn = {1537-5110},
  doi = {10.1016/j.biosystemseng.2023.08.010},
  author = {Frans P. Boogaard and Eldert J. {van Henten} and Gert Kootstra},
}

@article{rodriguez2024ground,
  title={A Ground Mobile Robot for Autonomous Terrestrial Laser Scanning-Based Field Phenotyping},
  author={Rodriguez-Sanchez, Javier and Johnsen, Kyle and Li, Changying},
  journal={arXiv preprint arXiv:2404.04404},
  year={2024}
}

@article{rodriguez2024fps,
author={Rodriguez-Sanchez, Javier  and Snider, John L.  and Johnsen, Kyle  and Li, Changying },
title={Cotton morphological traits tracking through spatiotemporal registration of terrestrial laser scanning time-series data},
journal={Frontiers in Plant Science},
volume={15},
year={2024},
doi={10.3389/fpls.2024.1436120},
}

@inproceedings{lobefaro2024iros,
author = {L. Lobefaro and M.V.R. Malladi and T. Guadagnino and C. Stachniss},
title = {{Spatio-Temporal Consistent Mapping of Growing Plants for Agricultural Robots in the Wild}},
booktitle = iros,
year = 2024,
pages={6375--6382},
}

@inproceedings{lobefaro2023iros,
author = {L. Lobefaro and M.V.R. Malladi and O. Vysotska and T. Guadagnino and C. Stachniss},
title = {{Estimating 4D Data Associations Towards Spatial-Temporal Mapping of Growing Plants for Agricultural Robots}},
booktitle = iros,
year = 2023,
pages={4212--4218},

}

@article{xiang2023isprs,
title = {A Review of panoptic segmentation for mobile mapping point clouds},
author = {Binbin Xiang and Yuanwen Yue and Torben Peters and Konrad Schindler},
journal = jprs,
volume = {203},
pages = {373-391},
year = {2023},
doi = {10.1016/j.isprsjprs.2023.08.008},
}

@article{genemola2021dib,
author = {Jordi Gené-Mola and Ricardo Sanz-Cortiella and Joan R. Rosell-Polo and Alexandre Escolà and Eduard Gregorio},
title = {PFuji-Size dataset: A collection of images and photogrammetry-derived 3D point clouds with ground truth annotations for Fuji apple detection and size estimation in field conditions},
journal = {Data in Brief},
volume = {39},
pages = {107629},
year = {2021},
doi = {10.1016/j.dib.2021.107629},
}

@article{schult2023icra,
  title     = {{Mask3D: Mask Transformer for 3D Semantic Instance Segmentation}},
  author    = {Schult, Jonas and Engelmann, Francis and Hermans, Alexander and Litany, Or and Tang, Siyu and Leibe, Bastian},
  journal = icra,
  year      = {2023}
}

@article{robert20243dv,
  title={Scalable 3D Panoptic Segmentation as Superpoint Graph Clustering},
  author={Robert, Damien and Raguet, Hugo and Landrieu, Loic},
  journal=threedv,
  year={2024}
}

@inproceedings{shin2024cvpr,
 author={Shin, Sangyun and Zhou, Kaichen and Vankadari, Madhu and Markham, Andrew and Trigoni, Niki},
 booktitle=cvpr,
 title={Spherical Mask: Coarse-to-Fine 3D Point Cloud Instance Segmentation with Spherical Representation},
 pages={4060--4069},
 year= {2024}
}

@article{xiao2025ijcv,
  author    = {Xiao, Zeqi and Zhang, Wenwei and Wang, Tai and Loy, Chen Change and Lin, Dahua and Pang, Jiangmiao},
  title     = {Position-Guided Point Cloud Panoptic Segmentation Transformer},
  journal   = ijcv,
  year      = {2025},
  volume    = {133},
  number    = {1},
  pages     = {275--290},
  doi       = {10.1007/s11263-024-02162-z}
}

@inproceedings{kolodiazhnyi2024cvpr,
  title={Oneformer3d: One transformer for unified point cloud segmentation},
  author={Kolodiazhnyi, Maxim and Vorontsova, Anna and Konushin, Anton and Rukhovich, Danila},
  booktitle=cvpr,
  pages={20943-20953},
  year={2024}
}

@inproceedings{hong2021cvpr,
    author    = {Hong, Fangzhou and Zhou, Hui and Zhu, Xinge and Li, Hongsheng and Liu, Ziwei},
    title     = {LiDAR-Based Panoptic Segmentation via Dynamic Shifting Network},
    booktitle = cvpr,
    year      = {2021},
    pages     = {13090-13099}
}

@article{rodriguez2020prl,
  title = {A computer vision system for automatic cherry beans detection on coffee trees},
  journal = {Pattern Recognition Letters},
  volume = {136},
  pages = {142-153},
  year = {2020},
  doi = {10.1016/j.patrec.2020.05.034},
  author = {Jhonn Pablo Rodríguez and David Camilo Corrales and Jean-Noël Aubertot and Juan Carlos Corrales},
}

@article{liu2024pr,
  title = {CS-net: Conv-simpleformer network for agricultural image segmentation},
  journal = {Pattern Recognition},
  volume = {147},
  pages = {110140},
  year = {2024},
  doi = {10.1016/j.patcog.2023.110140},
  author = {Lei Liu and Guorun Li and Yuefeng Du and Xiaoyu Li and Xiuheng Wu and Zhi Qiao and Tianyi Wang},
}

@article{chu2021prl,
title = {Deep learning-based apple detection using a suppression mask R-CNN},
journal = {Pattern Recognition Letters},
volume = {147},
pages = {206-211},
year = {2021},
doi = {10.1016/j.patrec.2021.04.022},
author = {Pengyu Chu and Zhaojian Li and Kyle Lammers and Renfu Lu and Xiaoming Liu},
}

@article{cardellicchio2024prl,
title = {Patch-based probabilistic identification of plant roots using convolutional neural networks},
journal = {Pattern Recognition Letters},
volume = {183},
pages = {125-132},
year = {2024},
doi = {10.1016/j.patrec.2024.05.010},
author = {A. Cardellicchio and F. Solimani and G. Dimauro and S. Summerer and V. Renò},
}

@article{zhu2024pr,
title = {Advancements in point cloud data augmentation for deep learning: A survey},
journal = {Pattern Recognition},
volume = {153},
pages = {110532},
year = {2024},
doi = {10.1016/j.patcog.2024.110532},
author = {Qinfeng Zhu and Lei Fan and Ningxin Weng},
}

@article{tang2024fsfs,
  author={Tang, Shixi  and Xia, Zilin  and Gu, Jinan  and Wang, Wenbo  and Huang, Zedong  and Zhang, Wenhao },
  title={High-precision Apple Recognition and Localization Method Based on {RGB-D} and Improved {SOLOv2} Instance Segmentation},
  journal={Frontiers in Sustainable Food Systems},
  volume={8},
  year={2024},
  doi={10.3389/fsufs.2024.1403872},
}

@article{wang2020nipsjournel,
  title={{SOLOv2}: Dynamic and Fast Instance Segmentation},
  author={Wang, Xinlong and Zhang, Rufeng and  Kong, Tao and Li, Lei and Shen, Chunhua},
  journal=nipsjournel,
  year={2020}
}

@article{ge2019ifac,
  title = {Instance Segmentation and Localization of Strawberries in Farm Conditions for Automatic Fruit Harvesting},
  journal = ifac,
  volume = {52},
  number = {30},
  pages = {294-299},
  year = {2019},
  doi = {10.1016/j.ifacol.2019.12.537},
  author = {Yuanyue Ge and Ya Xiong and P{{\textcircled{\textit{a}}}}l J. From},
}

@inproceedings{tan2019icml,
  title={{EfficientNet}: Rethinking Model Scaling for Convolutional Neural Networks},
  author={Tan, Mingxing and Le, Quoc},
  booktitle={International conference on machine learning},
  pages={6105--6114},
  year={2019},
}

@article{kang2020cea,
  title = {Fruit Detection, Segmentation and {3D} Visualisation of Environments in Apple Orchards},
  journal = cea,
  volume = {171},
  pages = {105302},
  year = {2020},
  doi = {10.1016/j.compag.2020.105302},
  author = {Hanwen Kang and Chao Chen},
}

\end{document}